# A Systematic Literature Review on Deep Learning-based Depth Estimation in Computer Vision

Ali Rohan[a,*], Md Junayed Hasan[a], Andrei Petrovski[a]

[a]*National Subsea Centre, School of Computing, Robert Gordon University, 3 International Ave, Dyce, Aberdeen, AB21 0BH, UK*

**Abstract**

Depth estimation (DE) provides spatial information about a scene and enables tasks such as 3D reconstruction, object detection, and scene understanding. Recently, there has been an increasing interest in using deep learning (DL)-based methods for DE. Traditional techniques rely on handcrafted features that often struggle to generalise to diverse scenes and require extensive manual tuning. However, DL models for DE can automatically extract relevant features from input data, adapt to various scene conditions, and generalise well to unseen environments. Numerous DL-based methods have been developed, making it necessary to survey and synthesize the state-of-the-art (SOTA). Previous reviews on DE have mainly focused on either monocular or stereo-based techniques, rather than comprehensively reviewing DE. Furthermore, to the best of our knowledge, there is no systematic literature review (SLR) that comprehensively focuses on DE. Therefore, this SLR study is being conducted. Initially, electronic databases were searched for relevant publications, resulting in 1284 publications. Using defined exclusion and quality criteria, 128 publications were shortlisted and further filtered to select 59 high-quality primary studies. These studies were analysed to extract data and answer defined research questions. Based on the results, DL methods were developed for mainly three different types of DE: monocular, stereo, and multi-view. 20 publicly available datasets were used to train, test, and evaluate DL models for DE, with KITTI, NYU Depth V2, and Make 3D being the most used datasets. 29 evaluation metrics were used to assess the performance of DE. 35 base models were reported in the primary studies, and the top five most-used base models were ResNet-50, ResNet-18, ResNet-101, U-Net, and VGG-16. Finally, the lack of ground truth data was among the most significant challenges reported by primary studies.

[*]Corresponding authors at National Subsea Centre, School of Computing, Robert Gordon University, Aberdeen AB10 7AQ, UK (A.R.) E-mail addresses: a.rohan@rgu.ac.uk; ali_rohan2003@hotmail.com (A.R.)

1. Introduction

Depth Estimation (DE) is an essential component of the field of computer vision, enabling machines to understand the spatial arrangement of a scene in images or videos. DE calculates the distance to each object in the image, similar to how humans perceive depth. This extra dimension of distance helps to create a 3-D representation of the environment, enhancing machines' abilities to navigate, interact, and comprehend visual information more efficiently [1]. DE has had a significant impact on the development of intelligent systems and human-machine interactions. It is now essential for applications such as autonomous vehicles [2] [3], robots [4], autonomous navigation [5] [6], obstacle avoidance [7], object detection [8], virtual reality, and augmented reality [9].

The techniques for DE have evolved rapidly over the past few decades. Early approaches relied heavily on depth cues like vanishing points [10] to infer depth and focus-defocus techniques [11] that used the variation in sharpness to estimate depth. However, these methods were not suitable for handling complex scenes, diverse environments, and non-constrained scenarios where these cues might not be applicable. Later, researchers developed several hand-crafted features-based methods like Scale-Invariant Feature Transform (SIFT) [12], which detects distinctive features in invariant images to scale variations, Conditional Random Field (CRF) [13], which uses probabilistic graphical models to capture contextual dependencies for structured prediction, Markov Random Field (MRF) [14], which uses stochastic models over a set of random variables and models spatial dependencies, and Speeded-up Robust Features (SURF) [15], which detects and describes interest points in images that are robust to scale and rotation changes. However, these methods had several limitations. For instance, SIFT has limited performance in highly textured and repetitive scenes [16], CRF is less efficient for large-scale problems [17], MRF requires careful modelling of pairwise potentials and struggles with large optimisation problems [14], and SURF struggles in cases of extreme scale and rotation changes, making it less robust to viewpoint variations [15].

On the contrary, researchers developed methods such as Stereo Block Matching (StereoBM) [18] and Semi-Global Block Matching (StereoSGBM) [19] to calculate disparities for DE. These methods match the defined area of blocks between two images taken from different perspectives. This breakthrough has laid the foundation for subsequent developments of traditional DE methods, especially in computer vision and robotics. Traditional DE methods are primarily based on stereo vision. These methods measure the binocular disparity between two images taken by cameras set at slightly different positions using stereo-matching algorithms and triangulation techniques. By adding spatial information, stereo-based DE provides accurate depth information, allowing one to understand the 3D structure of the scene and enhancing object recognition capabilities. However, stereo-based methods are considered delicate since minor miscalibrations in stereo camera systems and variations in scene illumination can lead to inaccurate and erroneous depth information [20]. Using binocular cameras makes it difficult to capture sufficient features for stereo matching when the scene has less or no texture [21].

Researchers have developed Monocular Depth Estimation (MDE) methods to overcome the limitations of stereo-based DE approaches. Unlike stereo-based methods, MDE uses only a single image captured by a camera to estimate depth. This makes it simple, cost-effective, and suitable for various devices and real-time applications. However, MDE is less accurate, especially in complex settings. Although MDE is an ill-posed problem to regress depth in 3D space due to the absence of a reliable stereoscopic relationship, deep learning techniques have considerably improved MDE by taking advantage of large-scale datasets and sophisticated neural network architectures [22]. The selection of the most suitable method for DE, whether stereo or monocular, is highly dependent on various variables such as scene complexity, available resources, and real-time processing requirements of the application.

Several review studies have been conducted in the recent past which focus on the development of deep learning for MDE and stereo-based DE. Reference [23] provided an overview of published papers between 2014 and 2020. Reference [24] reviewed MDE methods using different datasets and evaluation indicators. Reference [25] provided a comprehensive collection of outdoor DE techniques. Although recent review studies are more concentrated on deep learning-based MDE, there have been a few studies focusing on stereo-based DE methods. Reference [26] focused on the application of deep learning in stereo matching and disparity estimation. Reference [27] summarised the most used pipelines and discussed their benefits and limitations for stereo-based DE.

The studies mentioned above only focus on either monocular or stereo-based DE, without considering the wider panorama of this field. However, reviewing depth estimation comprehensively, rather than focusing solely on specific modalities, like monocular or stereo methods, would offer a holistic understanding of the field's trends, challenges, and integration potential. An overview and analysis of the state-of-the-art are essential, not only for individual methodologies but also for their overall impact. It is crucial to systematically examine the current advancements in both monocular and stereo approaches. Therefore, this review aims to provide a comprehensive overview of recent developments in deep learning-based DE in computer vision. It will cover the importance and critical applications of DE in computer vision, and the different datasets available, including details about the quantity, quality, and type of data used in the literature. Furthermore, it will discuss the performance analysis of DE approaches, challenges reported

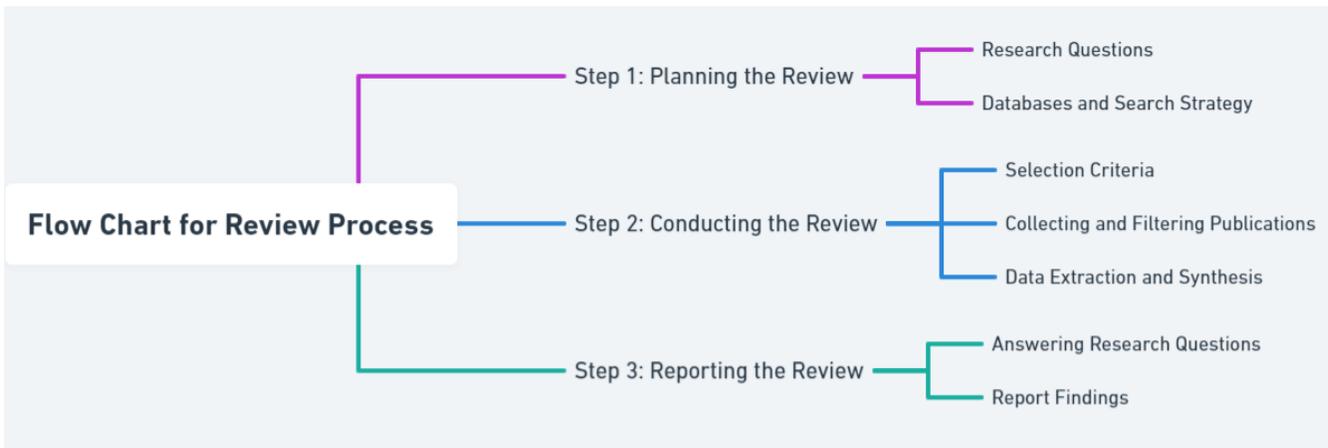

Figure 1: Flowchart of the review process.

in the literature, potential solutions, and future directions.

The details of this study are presented in the following sections. Section 2 provides the details of the methodology used to conduct this review. Section 3 provides the results answering each research question. Lastly, section 4 presents the conclusions.

## 2. Methodology

*2.1. Review Protocol*

This study adheres to the "Guidelines for Performing Systematic Literature Reviews in Software Engineering" by [28] The process of Systematic Literature Review (SLR) involves three primary steps, as illustrated in Figure 1.

The initial step involves planning the review, which includes identifying the review's necessity, formulating research questions, and developing search strategies. These strategies concentrate on selecting relevant databases, creating effective search strings, and defining selection criteria. The second step is executing the review, which involves choosing primary studies and extracting and synthesising data. Search strings are utilized to scrutinise titles, abstracts, and keyword fields in selected databases. Publications that meet the selection criteria are shortlisted, and a quality assessment process ensures that only high-quality publications are included. This step involves extracting and synthesizing data required to address research questions. The final step is reporting the findings. Research question answers and outcomes are presented through supporting figures and tables, culminating in a comprehensive representation of the study's results.

*2.2. Research Questions*

This study posits five critical research questions aimed at exploring, collecting, and presenting recent trends, challenges, and the importance of the application of deep learning-based methods in DE.

**RQ.1:** What are the importance and the critical applications of DE in computer vision?

**RQ.2:** What are the recent advancements and developments in monocular and stereo DE approaches?

**RQ.3:** What are the available datasets for DE, and what are their type, quantity, and quality?

**RQ.4:** What methodologies and metrics are used to assess the outcome of DE?

**RQ.5:** What are the main challenges and limitations of DE?

*2.3. Database and Search Strategy*

For this SLR study, seven popular databases were selected to search for related publications. These databases were Google Scholar, IEEE Xplore, ScienceDirect, SpringerLink, Scopus, Web of Science, and Wiley. Relevant keywords were combined into search strings, including "depth estimation" and "deep learning" as initial keywords, and synonymous terms such as "depth perception", "depth mapping", "monocular", "stereo", and "multi-view" related to "depth estimation", as well as "artificial intelligence" and "machine learning" related to "deep learning". The general search string used was ("deep learning" OR "machine learning" OR "artificial intelligence") AND ("depth mapping" OR "depth estimation" OR "depth perception") AND ("monocular" OR "stereo" OR "multi-view"). The general string was then modified based on each database's specific search criteria. For example, the keywords were adjusted to account for character limits for Google Scholar and restrictions on Boolean characters for ScienceDirect. Each database's abstract, title, and keyword fields were searched, except for Wiley and SpringerLink, which only allowed keyword searches within the publications themselves. Filtering was done using selection and quality criteria. In total, 1284 publications were found through the search of the selected databases. The specific search strings used for each database are listed below:

Google Scholar: [("computer vision" AND "depth estimation")] AND [("deep learning" OR "machine learning" OR "artificial intelligence") AND ("monocular" OR "stereo" OR "Multiview")]

IEEE Xplore: ("deep learning" OR "machine learning" OR "artificial intelligence") AND ("computer vision" OR "depth estimation" OR "depth perception" OR "depth mapping") AND ("monocular" OR "stereo" OR multiview)

ScienceDirect: ("deep learning" OR "machine learning" OR "artificial intelligence") AND ("computer vision" OR "depth estimation" OR "depth perception") AND ("monocular" OR "stereo" OR multiview)

Scopus: TITLE-ABS-KEY ( "deep learning" OR "machine learning" OR "artificial intelligence" AND "computer vision" OR "depth estimation" OR "depth perception" OR "depth mapping" AND "monocular" OR "stereo" OR "multiview" ) AND ( LIMIT-TO ( OA, "all" ) ) AND ( LIMIT-TO ( PUBYEAR, 2023 ) OR LIMIT-TO ( PUBYEAR, 2022 ) OR LIMIT-TO ( PUBYEAR, 2021 ) OR LIMIT-TO ( PUBYEAR, 2020 ) OR LIMIT-TO ( PUBYEAR, 2019 ) OR LIMIT-TO ( PUBYEAR, 2018 ) ) AND ( LIMIT-TO ( DOCTYPE, "ar" ) )

SpringerLink, Wiley: ("deep learning" OR "machine learning" OR "artificial intelligence") AND ("computer vision" OR "depth estimation" OR depth perception OR "depth mapping" ) AND ("monocular" OR "stereo" OR "multiview")(anywhere)

Web of Science: TI=(("deep learning" OR "machine learning" OR "artificial intelligence") AND ("computer vision" OR "depth estimation" OR depth perception OR "depth mapping" ) AND ("monocular" OR "stereo" OR "multiview")) OR AB=(("deep learning" OR "machine learning" OR "artificial intelligence") AND ("computer vision" OR "depth estimation" OR depth perception OR "depth mapping" ) AND ("monocular" OR "stereo" OR "multiview")) OR AK=(("deep learning" OR "machine learning" OR "artificial intelligence") AND ("computer vision" OR "depth estimation" OR depth perception OR "depth mapping" ) AND ("monocular" OR "stereo" OR "multiview"))

*2.4. Selection Criteria*

During the initial search, numerous publications were discovered; however, a significant number of them were irrelevant to the subject of DE. To address this issue and to improve the quality, specific criteria were established to filter out irrelevant publications. Each publication was evaluated against these criteria, and only those that did not violate any of the exclusion criteria were retained. This methodology was based on the guidelines outlined by [28]. To ensure agreement on the eligibility of each publication, the Cohen Kappa statistic [29] was utilized. Finally, out of the 1284 publications, only 128 were deemed relevant and included in further study. The exclusion criteria used are given below:

1. The publication is not related to DE in computer vision.
2. The publication either contains duplicated content or has been retrieved from another database.

3. The publication is not written in English, open access, or the full text of the study is not available.

4. The publication is categorized as a book chapter, review, conference abstract, overview, data article, report, survey, mini-review, short communication, or comparative study.

5. The publication is either a pre-print or has not been peer-reviewed.

6. The publication was published before the year 2018.

*2.5. Collecting and Filtering Publications*

The 128 publications underwent a rigorous evaluation to ensure that only high-quality primary studies were included. To evaluate the publications, we utilized quality assessment criteria derived from the study by [30]. Each publication was assessed based on a series of questions, with a score of 1 (yes), 0 (no), or 0.5 (maybe) assigned to each response. The total score for each publication was then calculated, and any publications scoring below three were excluded. Finally, 59 publications were selected as primary studies. Figure 2 shows the overall process for the selection of primary studies. The quality criteria employed in this study are given below:

1. Are the study's aims and objectives clearly stated?
2. Is the study's scope, methodology, and experimental design clearly defined?
3. Is the research process documented appropriately?
4. Have all of the study questions been answered?
5. Have any negative findings been presented?
6. Do the study's conclusions align with its goals and purpose?

*2.6. Data Extraction and Synthesis*

A comprehensive review of 59 primary studies was conducted to extract pertinent data for each research question. The specifics of the selected primary studies for this SLR study can be found in Table 1. Using a Microsoft Excel spreadsheet, the relevant data was organized in rows for each primary study and columns for each research question. The extracted data was focused on addressing the research questions, including the objectives, the significance and practical applications of DE, the current global research trends for DE, the latest advancements and developments in monocular and stereo DE approaches, available datasets for DE, their type, quantity, and quality, performance evaluation methods and metrics, the publication year and journal details, and challenges associated with the application of DL for DE. After synthesising the extracted data, each research question was thoroughly answered. The results of this SLR study are detailed in the following section.

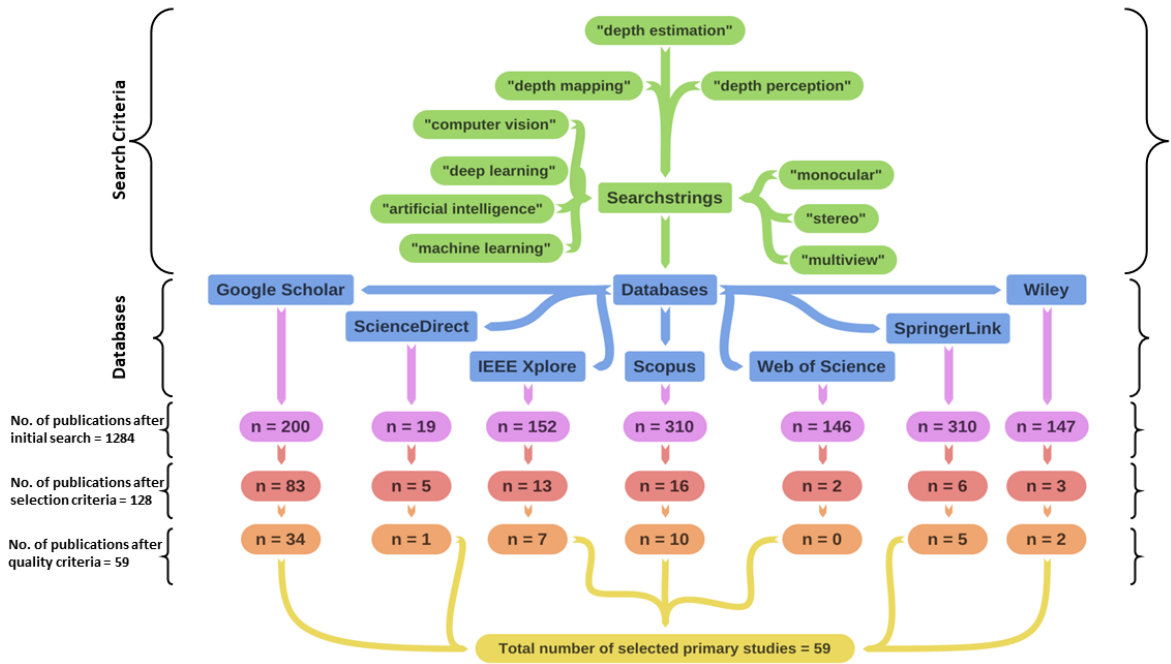

Figure 2: The overall process of selection of the primary studies.

## 3. Results

*3.1. RQ.1: Importance and the critical applications of DE*

In 59 primary studies, researchers have identified the numerous uses of DE in various applications. Figure 3 provides an insightful overview of these applications, ranked from the most to least common. The six most common applications of DE include autonomous vehicles, robotics, 3D scene reconstruction, Augmented Reality (AR), scene understanding, and Simultaneous Localisation and Mapping (SLAM). These applications also encompass autonomous navigation, Virtual Reality (VR), and Unmanned Aerial Vehicles (UAVs), among others.

In autonomous vehicles, DE enables accurate measurement of the distance of objects, allowing vehicles to recognize and navigate around obstacles. This information is used to create a 3D map of the environment, enabling the vehicle to make real-time decisions and navigate safely in diverse and challenging conditions. In robotics, DE enhances the versatility and efficiency of robotic systems across various domains. Robots leverage DE for object manipulation, ensuring precise interactions in industrial and manufacturing settings. In surveillance, depth perception enhances object tracking and anomaly detection. Human-robot interaction benefits from DE, enabling robots to understand and respond to human gestures and movements. Additionally, depth information aids in 3D mapping, crucial for mapping unknown

Table 1: Details of the selected primary studies.

| No. | Source | Article Title | Reference | Abbreviations |
|---|---|---|---|---|
| 1 | | Progressive Hard-Mining Network for Monocular Depth Estimation | [31] | PHN |
| 2 | | Monocular depth estimation with hierarchical fusion of dilated CNNs and soft-weighted-sum inference | [32] | Deep CNN |
| 3 | | T2Net: Synthetic-to-Realistic Translation for Solving Single-Image Depth Estimation Tasks | [33] | T2Net |
| 4 | | AdaDepth: Unsupervised Content Congruent Adaptation for Depth Estimation | [34] | AdaDepth |
| 5 | | Single View Stereo Matching | [35] | SVSM |
| 6 | | Unsupervised Learning of Monocular Depth Estimation and Visual Odometry with Deep Feature Reconstruction | [36] | ConvNetD |
| 7 | | 3D Packing for Self-Supervised Monocular Depth Estimation | [37] | PackNet |
| 8 | | UnOS: Unified Unsupervised Optical-Flow and Stereo-Depth Estimation by Watching Videos | [38] | UnOS |
| 9 | | Learn Stereo, Infer Mono: Siamese Networks for Self-Supervised, Monocular, Depth Estimation | [39] | Deep Siamese |
| 10 | | Single Image Depth Estimation Trained via Depth From Defocus Cues | [40] | DDC |
| 11 | | Learning Monocular Depth Estimation Infusing Traditional Stereo Knowledge | [41] | monoResMatch |
| 12 | | Recurrent Neural Network for (Un-)Supervised Learning of Monocular Video Visual Odometry and Depth | [42] | RNN for Depth |
| 13 | | Spatial Correspondence With Generative Adversarial Network: Learning Depth From Monocular Videos | [43] | SC-GAN |
| 14 | | Self-Supervised Monocular Trained Depth Estimation Using Self-Attention and Discrete Disparity Volume | [44] | SADDV-MDE |
| 15 | | AdaBins: Depth Estimation Using Adaptive Bins | [45] | AdaBins |
| 16 | | Mixed-Scale Unet Based on Dense Atrous Pyramid for Monocular Depth Estimation | [46] | MAPUnet |
| 17 | Google Scholar | D-Net: A Generalised and Optimised Deep Network for Monocular Depth Estimation | [47] | D-Net |
| 18 | | Revisiting Stereo Depth Estimation From a Sequence-to-Sequence Perspective with Transformers | [48] | STTR |
| 19 | | MonoIndoor: Towards Good Practice of Self-Supervised Monocular Depth Estimation for Indoor Environments | [49] | MonoIndoor |
| 20 | | Boosting Monocular Depth Estimation with Lightweight 3D Point Fusion | [50] | Point-Fusion |
| 21 | | Self-supervised monocular depth estimation with direct methods | [51] | MDE-DM |
| 22 | | Monocular Depth Estimation Using Information Exchange Network | [52] | IEN |
| 23 | | Attention-based context aggregation network for monocular depth estimation | [53] | ACAN |
| 24 | | Efficient unsupervised monocular depth estimation using attention guided generative adversarial network | [54] | AG-GAN |
| 25 | | A lightweight network for monocular depth estimation with decoupled body and edge supervision | [55] | DBES-MDE |
| 26 | | Self-supervised Monocular Depth Estimation for All Day Images using Domain Separation | [56] | DS-MDE |
| 27 | | Revealing the Reciprocal Relations between Self-Supervised Stereo and Monocular Depth Estimation | [57] | StereoNet /SingleNet |
| 28 | | Swin-Depth: Using Transformers and Multi-Scale Fusion for Monocular-Based Depth Estimation | [58] | Swin-Depth |
| 29 | | Neural Window Fully-connected CRFs for Monocular Depth Estimation | [59] | FC-CRFs |
| 30 | | H-Net: Unsupervised Attention-based Stereo Depth Estimation Leveraging Epipolar Geometry | [60] | H-Net |
| 31 | | Unsupervised Monocular Depth Estimation Using Attention and Multi-Warp Reconstruction | [61] | AMWR |
| 32 | | Deep Learning-Based Incorporation of Planar Constraints for Robust Stereo Depth Estimation in Autonomous Vehicle Applications | [62] | PCStereo |
| 33 | | CORNet: Context-Based Ordinal Regression Network for Monocular Depth Estimation | [63] | CORNet |
| 34 | | Self-supervised Monocular Depth Estimation Using Hybrid Transformer Encoder | [64] | THE |
| 35 | | LW-Net: A Lightweight Network for Monocular Depth Estimation | [65] | LW-Net |
| 36 | | Leveraging Contextual Information for Monocular Depth Estimation | [66] | CI-MDE |
| 37 | IEEE Xplore | Joint Attention Mechanisms for Monocular Depth Estimation With Multi-Scale Convolutions and Adaptive Weight Adjustment | [67] | JAM-MDE |
| 38 | | Efficient and High-Quality Monocular Depth Estimation via Gated Multi-Scale Network | [68] | GMSN |
| 39 | | Attention-Based Dense Decoding Network for Monocular Depth Estimation | [69] | DDN-MDE |
| 40 | | Monocular Depth Estimation Based on Multi-Scale Graph Convolution Networks | [70] | Multiscale-GCN |
| 41 | | Monocular Depth Estimation Based on Multi-Scale Depth Map Fusion | [71] | DFFN |
| 42 | Science Direct | SABV-Depth: A biologically inspired deep learning network for monocular depth estimation | [72] | SABV-Depth |
| 43 | | Semi-Supervised Adversarial Monocular Depth Estimation | [73] | AMDE. |
| 44 | | MiniNet: An extremely lightweight convolutional neural network for real-time unsupervised monocular depth estimation | [74] | MiniNet |
| 45 | | PVStereo: Pyramid Voting Module for End-to-End Self-Supervised Stereo Matching | [75] | PVStereo |
| 46 | Scopus | ADAADepth: Adapting Data Augmentation and Attention for Self-Supervised Monocular Depth Estimation | [76] | ADAADepth |
| 47 | | Parallax attention stereo matching network based on the improved group-wise correlation stereo network | [77] | PA-Net |
| 48 | | SVDistNet: Self-Supervised Near-Field Distance Estimation on Surround View Fisheye Cameras | [78] | SVDistNet |
| 49 | | Detaching and Boosting: Dual Engine for Scale-Invariant Self-Supervised Monocular Depth Estimation | [79] | SI-MDE |
| 50 | | Visual Attention-Based Self-Supervised Absolute Depth Estimation Using Geometric Priors in Autonomous Driving | [80] | VADepth |
| 51 | | Depth Estimation Based on Monocular Camera Sensors in Autonomous Vehicles: A Self-supervised Learning Approach | [3] | DE-MCS |
| 52 | | Depth estimation for advancing intelligent transport systems based on self-improving pyramid stereo network | [81] | SIPSNet |
| 53 | | Digging Into Self-Supervised Monocular Depth Estimation | [82] | Monodepth2 |
| 54 | | Vision Transformers for Dense Prediction | [83] | VT |
| 55 | Springer Link | HITNet: Hierarchical Iterative Tile Refinement Network for Real-time Stereo Matching | [84] | HITNet |
| 56 | | RAFT-Stereo: Multilevel Recurrent Field Transforms for Stereo Matching | [85] | RAFT-Stereo |
| 57 | | Practical Stereo Matching via Cascaded Recurrent Network with Adaptive Correlation | [86] | CREStereo |
| 58 | Wiley | Semantically guided self-supervised monocular depth estimation | [87] | SG-MDE |
| 59 | | Self-supervised monocular depth estimation via asymmetric convolution block | [88] | ACB |

environments and creating detailed spatial representations for robots to navigate effectively in diverse and dynamic surroundings. DE is also instrumental in advancing the fidelity and utility of 3D scene reconstruction across diverse disciplines. Medical imaging benefits from precise DE for reconstructing anatomical structures in three dimensions, aiding in diagnostics and surgical planning. Archaeology and cultural heritage preservation leverage DE for creating detailed 3D models of artefacts and historical sites. In AR, DE enhances realistic virtual object placement within the real world. Accurate depth information ensures proper occlusion and alignment, creating immersive AR experiences. This technology is integral for applications like virtual try-ons in e-commerce, interactive gaming, and architectural visualization, enhancing user engagement and interaction in AR environments. DE is vital for scene understanding in computer vision. It enables precise object recognition, segmentation, and spatial awareness, ensuring a comprehensive understanding of scenes for improved decision-making and interactions. In SLAM, DE enables robots and autonomous systems to construct detailed 2D and 3D maps of their surroundings while simultaneously tracking their position, facilitating navigation in unknown environments, and enhancing overall system performance. Furthermore, DE has versatile applications that underscore its significance in advancing technologies related to spatial perception and interaction. It is used for autonomous navigation, obstacle detection, and path planning in vehicles. It enhances the immersive quality of VR by accurately situating virtual objects in three-dimensional space. Semantic segmentation and scene recognition become more detailed and context-aware with DE. It also contributes to accurate pose estimation and improves the recognition of human actions in videos.

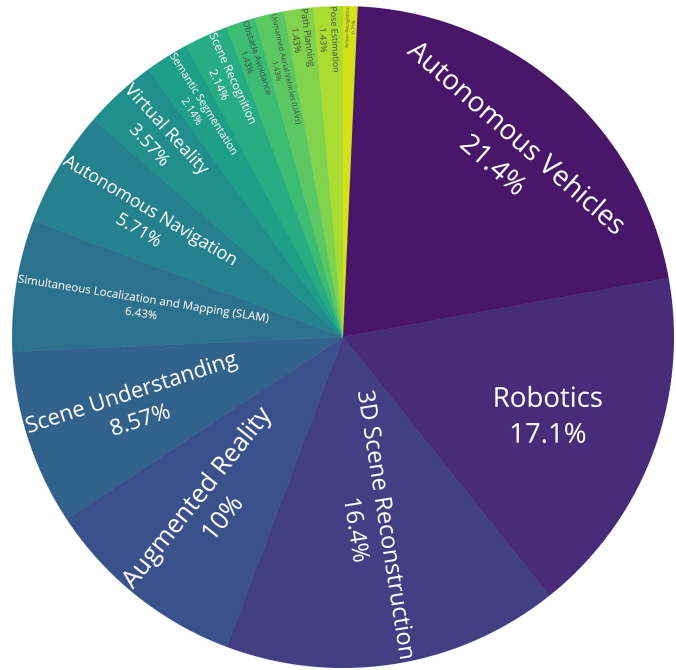

Figure 3: Depth estimation and its applications.

*3.1.1. Primary studies and global research trend for DE*

Figure 4 presents the breakdown of primary studies based on their year of publication and the journal they were published. This SLR study focuses on research studies that were published within the last five years, covering the years from 2018 to 2023. As expected, the number of publications has steadily increased from 2018 to 2021, with the surge in popularity of deep learning-based methods being the primary driving factor. The year 2021 had the highest number of related publications. These publications were featured in a total of 24 different journals, with Computer Vision and Pattern Recognition (CVPR) leading the way, followed by IEEE Access and the International Conference on Computer Vision (ICCV), with 15, 9, and 8 publications respectively, from 2018 to 2023. Figure 5 showcases the current global research trend for DE using DL through a Sankey chart. The chart is segmented into three categories: countries with the highest number of related primary studies, learning paradigms, and depth estimation types. Based on the data, China has taken the lead in DL for DE research with 57.19% of the publications, followed by the USA at 16.84%, Australia at 5.26%, Germany at 4.21%, South Korea at 2.81%, and the UK at 2.81% among others. The deep learning methods involve a variety of ways for algorithms to acquire knowledge, including supervised, self-supervised, semi-supervised, and unsupervised learning paradigms. Primary studies have reported the use of all four paradigms. Supervised learning entails training a model on a labelled dataset of input-output pairs. This enables the algorithm to learn how to map inputs to corresponding desired outputs, facilitating accurate predictions on new data. Of the primary studies, 41.75% focused on the use of supervised learning. Self-supervised learning involves the model generating its labels from input data, promoting unsupervised learning through pretext tasks. For example, the model may predict missing parts of an image, leading to improved feature representation. Of the primary studies, 34.39% focused on the use of self-supervised learning. It is interesting to note that these studies are more recent, suggesting a shift towards the self-supervised learning paradigm in recent years. Unsupervised learning involves training models on unlabelled data to uncover inherent patterns, structures, or representations. This approach enables tasks like clustering, dimensionality reduction, and generative modelling without explicit guidance. Of the primary studies, 14.74% focused on unsupervised learning. On the contrary, semi-supervised learning uses both labelled and unlabelled data during training. This approach leverages both types of information to enhance model performance, making it particularly useful when labelled data is scarce or difficult to obtain. Of the primary studies, 9.12% focused on the use of semi-supervised learning.

As noted in the manuscript introduction, depth estimation can be divided into two primary types: stereo and monocular. Nevertheless, recent research has explored the use of Multiview DE, which involves a 360° perception of the scene geometry. Stereo DE uses two synchronised cameras to capture a scene from different perspectives, calculating depth information by using the disparity between corresponding points. This method provides highly accurate 3D perception. Conversely, MDE uses a single camera to estimate depth from a 2D image. Multiview DE combines data from multiple cameras or viewpoints to enhance depth perception by incorporating information from different angles, aiming to improve the accuracy of depth estimates, particularly in complex scenes. Of the primary studies reviewed, 22.11% focused on stereo-based DE, 75.44% focused on MDE, and only 2.46% on Multiview DE. Although stereo-based DE provides accurate depth estimation, it is sensitive to calibration errors, and feature matching becomes less reliable in low-texture or textureless regions, affecting the overall robustness of depth calculations. The study shows that current research trends are more focused on MDE, which often utilizes neural networks, making it cost-effective and widely applicable. However, MDE has inherent limitations in accurately perceiving 3D structures due to the inherent ambiguity of depth from a single 2D image. It struggles with depth variations, occlusions, and scale ambiguities, impacting the precision of depth predictions.

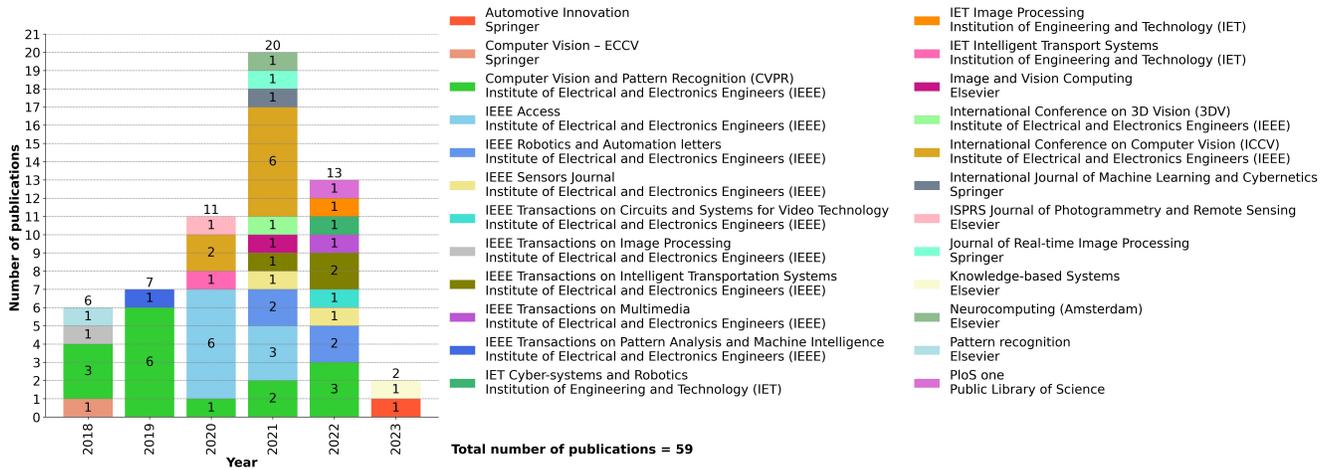

Figure 4: The breakdown of primary studies.

3.2. RQ.2: Recent advancements and developments in monocular and stereo DE approaches

In recent times, attention mechanisms such as self-attention and channel attention have been incorporated into DL-based MDE and stereo DE to capture contextual information more effectively. These mechanisms enable networks to focus on relevant features, which enhances the accuracy of DE [89]. Furthermore, unsupervised learning strategies are being leveraged in MDE, which eliminates the need for explicit ground-truth depth data. By using unlabelled monocular video sequences, these approaches enhance depth predictions through improved pose prediction and attention mechanisms [90]. Contemporary MDE networks utilize hierarchical feature extraction, combining low-resolution features that capture long-range context with fine-grained features describing local context. This approach has been reported to enhance the representation of specific semantics and improve overall feature extraction [91]. Additionally, recent models focus on designing efficient frameworks to address computational challenges. This includes leveraging lightweight encoder-decoder structures, introducing novel attention architectures, and incorporating factorized convolutions to reduce the number of model parameters and enhance computa-tional efficiency [92]. Several novel approaches have been introduced to improve scalability across different scenes and scales. This includes adaptive multi-scale convolutions, dynamic cross-attention modules, and scale-sensitive features to handle variations in object size and scene complexity [93]. Monocular DE models incorporate multi-task learning strategies, simultaneously predicting additional semantic information. This leads to more robust and interpretable depth maps, aiding in scene understanding [94].

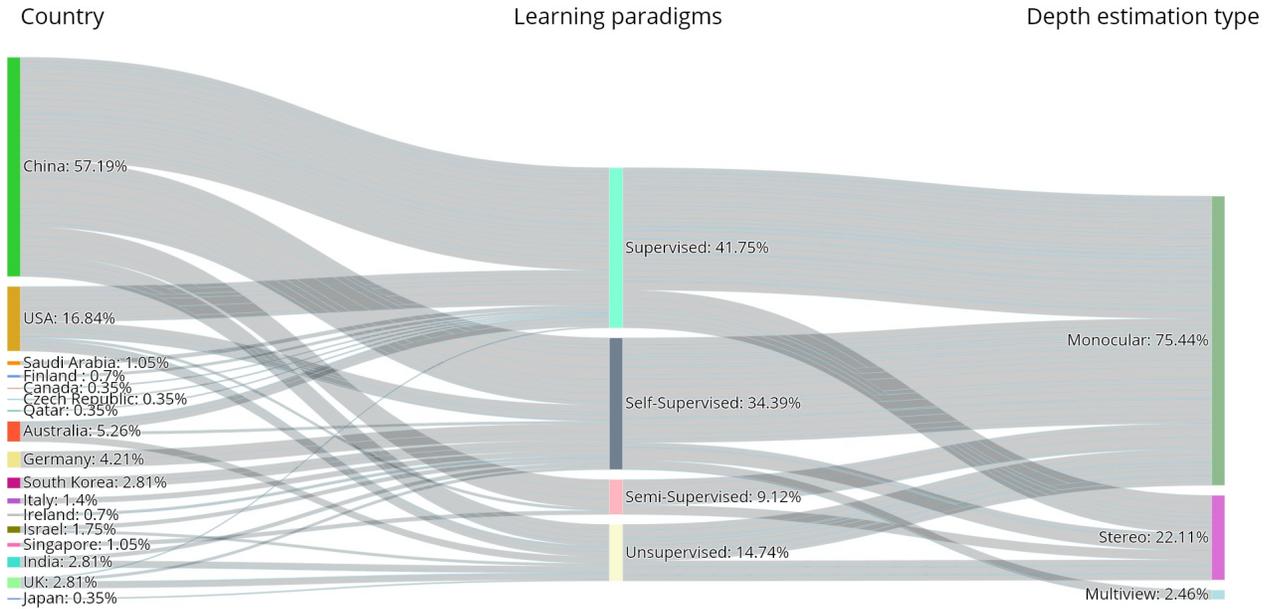

Figure 5: Current global research trend for depth estimation.

On the contrary, recent advancements in stereo DE focus on end-to-end learning, where deep neural networks are trained to predict depth maps directly from stereo image pairs. This approach eliminates the need for handcrafted features and enhances the model's ability to capture intricate depth details [95]. Stereo model refinement has been reported to refine disparity estimations using advanced cost volume processing. Techniques such as pyramid voting modules and recurrent refinement units are reported to contribute to more accurate and reliable disparity maps, especially in challenging scenarios [75]. Additionally, the integration of semantic information into stereo DE has become a prominent trend. Models leverage semantic cues to enhance stereo correspondences and optimal transport algorithms are employed to suppress attention in non-visible areas, improving disparity estimation [96]. Adaptive fusion techniques, such as occlusion-aware distillation modules and occlusion-aware fusion modules, have been introduced to fuse depth predictions from different sources. These methods improve the reliability of depth estimates, especially in regions with occlusions or challenging lighting conditions [57]. Stereo DE models now explore ordinal regression techniques, predicting depth maps in an ordinal manner [97]. Context-based encoders, boundary enhancement modules, and feature optimization modules contribute to the refinement of multi-scale features for more accurate depth predictions [98]. The recent advancements in both MDE and SDE showcase a paradigm shift towards more efficient, accurate, and context-aware approaches. These developments not only improve the fundamental understanding of scenes in various applications but also contribute to the broader goals of the application of deep learning-based methods in computer vision.

*3.3. Datasets for DE, their type, quantity, and quality*

The use of data is critical to the advancement of DL, as it lays the foundation for training and evaluating DL models. The accuracy and robustness of these models are directly impacted by the quantity, quality and type of datasets used. With comprehensive and diverse datasets, deep learning models can generalize effectively and perform well across various scenarios. In 59 primary studies, researchers employed 20 publicly available datasets to train, test, and evaluate deep learning models for DE. These datasets, which were recorded under different environmental settings such as outdoor, indoor, and synthetic, are available online. Table 2 provides details on the datasets used for DE, including their type, quantity, and quality. Of the 20 datasets, five were recorded outdoors, seven were recorded indoors, two were recorded in both indoor and outdoor environments, and six were recorded in synthetic environments. The number of images or videos used to record these datasets varied, and no established guidelines were found in primary studies on the amount of data needed to create tangible DL models. The Make3D dataset had the least amount of data, with only 534 RGB images and corresponding depth maps, and included both indoor and outdoor scenes. The Oxford RobotCar dataset had the most data, with 20 million RGB images with LIDAR scans and included solely outdoor scenes. Most of the datasets were recorded for RGB images with either corresponding depth maps or LIDAR scans as ground truths. For quality, the Lightfield dataset had the lowest image resolution of 512 x 512 pixels, while the DDAD dataset had the highest resolution of 4032 x 6048 pixels.

The usage rate of public datasets in primary studies was analysed, and Figure 6 depicts the results. Among the available datasets, the KITTI dataset emerged as the most used one with a usage rate of 45.30%. This dataset provides high-resolution images, lidar point clouds, and radar data, spanning various urban and highway environments. The dataset's diversity offers different testing scenarios, making it an invaluable resource for evaluating algorithms. Moreover, it features ground truth annotations making it an essential benchmark for autonomous driving technology research. NYU Depth V2 is the second most used dataset with a usage rate of 20.51%. It comprises indoor scenes captured by Microsoft Kinect, providing RGB images and corresponding depth maps. This dataset's diversity in indoor environments covering residential and office spaces makes it crucial for depth estimation, 3D scene understanding, and other computer vision applications. Cityscapes, with a usage rate of 6.84%, is the third most used dataset, consisting of high-quality images captured across diverse urban landscapes. It provides rich annotations, including semantic segmentation and instance-level annotations for various object classes, making it a crucial resource for urban scene understanding. Make3D, with a usage rate of 10.26%, is the fourth most used dataset. This dataset includes indoor and outdoor scenes with corresponding high-resolution images and accurate depth annotations. Both ETH3D and Middlebury dataset V3 were reported by only 2.56% of the primary studies among the other least used datasets.

Table 2: Details on the datasets used for depth estimation.

| Datasets | Environment | Data Quantity | Data Type | Data Quality (h x w) | Reference |
|---|---|---|---|---|---|
| CityScapes | Outdoor | 25,000 Images | RGB images | 1024 x 2048 px | [99] |
| DDAD | Outdoor | 102,680 Images | RGB images, LiDAR scans | 1216 x 1936 px | [37] |
| ETH3D | Indoor & Outdoor | 25 Scenes/ Videos | RGB images, LiDAR scans | 4032 x 6048 px | [100] |
| EURoCMAV | Indoor | 11 Scenes/ Videos | RGB Images 3D scans | 1365 x 2048 px | [101] |
| Falling Things | Synthetic | 61,500 Images | RGB-D images | 540 x 960 px | [102] |
| Flying things 3D | Synthetic | 25,000 Images | RGB-D images | 540 x 960 px | [103] |
| HKUST-Drive | Synthetic | 11,568 Images | RGB-D images | 1080 x 1920 | [75] |
| InStereo2K | Indoor | 2,050 Images | RGB-D images | 960 x 1280 px | [104] |
| KITTI | Outdoor | 93,000 Images | RGB images, LiDAR scans | 370 x 1224 px | [105] |
| Lightfield | Synthetic | 3,443 Images | RGB-D images | 512 x 512 px | [106] |
| Make3D | Indoor & Outdoor | 534 Images | RGB-D images | 1704×2272 px | [107] |
| MatterPort3D | Indoor | 194,400 Images | RGB-D images | 1024 x 1280 px | [108] |
| Middlebury dataset V3 | Indoor | 33 datasets | RGB-D images | 1988 x 2964 px | [109] |
| NuScenes | Outdoor | 40,157 Images | RGB images, LiDAR scans | 900 x 1600 px | [110] |
| NYU Depth v2 | Indoor | 464 Scenes/ Videos | RGB-D images | 480 x 640 px | [111] |
| Oxford RobotCar | Outdoor | 20 million Images | RGB images, LiDAR scans | 1024 x 1024 px | [112] |
| Physically-Based Rendering Dataset | Synthetic | 568,793 Images | RGB-D images | 480 x 640 px | [113] |
| RGBD | Indoor | 22 Scenes/ Videos | RGB-D images | 480 x 640 px | [114] |
| Sintel | Synthetic | 1,064 Images | RGB-D images | 436 x 1024 px | [115] |
| SUN RGB-D | Indoor | 10,335 Images | RGB-D images | 480 x 640 px | [116] |

Table 3, 4, and 5 provide a comprehensive summary of primary studies based on training data for DE. Specifically, this table includes information regarding the datasets utilised in training, the methods of training supervision, the amount of training data, the image size, and the base models used for the development of DE models.

It is noteworthy that most studies have employed multiple datasets for training, testing, and evaluation. The dataset used for testing and evaluation was distinct from the one used for training, a practice that enabled the assessment of the model's ability to generalise to new and unseen data, as well as the evaluation of the model's performance across diverse scenarios and datasets.

Table 3: Summary of primary studies based on training data for depth estimation.

| Methods | Datasets | Training Supervision | Training Data Quantity | Image Size | Base Model |
|---|---|---|---|---|---|
| PHN | NYU Depth v2, KITTI, Make3D | Stereo | 12,000 images, 4,000 images, 534 images | 240 x 320, N/A, 460 x 345 | VGG-16, ResNet-18, ResNet-50 |
| Deep CNN | NYU Depth v2, KITTI, Make3D | N/A | 12,000 images, 22,600 images, 534 images | 240 x 320, 188 x 620, 460 x 345 | ResNet-50, ResNet-101, ResNet-152 |
| T2Net | KITTI, Make3D | N/A | 22,600 images, 400 images | N/A | ResNet, PatchGAN, VGG |
| AdaDepth | NYU Depth v2, KITTI, Physically-Based Rendering Dataset | Stereo | 1449 images, 22,600 images, 100,000 images | 228 x 304, 256 x 512, N/A | ResNet-50 |
| SVSM | KITTI | Stereo | 22,600 images | 640 x 192 | VGG-16, DispNetC [26], DipFulNet [28] |
| ConvNetD | KITTI | Stereo | 22,600 images | 608 x 160 | ResNet-50 |
| PackNet | KITTI, DDAD, NuScenes, CityScapes | Monocular | 39,810 images, 17,050 images, 6019 images, 88,250 images | 640 x 192, N/A, N/A, N/A | N/A |
| UnOS | KITTI | Stereo | 200 sequences | 832 x 256 | N/A |
| Deep Siamese | KITTI | Stereo | 22,600 images | N/A | DispNet[41], ResNet, VGG |
| DDC | Lightfield, KITTI, NYU Depth v2, Make3D | Depth from de-focus cues | 3,143 images, 22,600 images, N/A, 400 images | 460 x 345, N/A, N/A, N/A | DeepLabV3+ [4, 5], ResNet |
| monoResMatch | KITTI, Cityscapes | Stereo | 22,600 images, N/A | N/A | N/A |
| RNN for Depth | KITTI | Multi-view | 45,200 10-frame sequences | $128 \times 416$ | RNN, ConvLSTM, VGG-16 |
| SC-GAN | NYU Depth v2, Make3D, KITTI, Cityscapes | N/A | 249 scenes, 400 images, 22,000 images, 20,000 images | 304 x 208, 320 x 240, N/A, N/A | PatchGAN |
| SADDV-MDE | KITTI, Cityscapes | Monocular | 22,600 images, 150,000 images | N/A | ResNet-50, PatchGAN |
| AdaBins | KITTI, Make3D | Monocular + Stereo | 22,600 images | N/A | U-net [53], ResNet-18 |
| MAPUnet | KITTI, Make3D | Monocular | 22,600 images | N/A | U-net, ResNet |
| D-Net | KITTI, NYU Depth v2 | Monocular | 22,600 images, 24,231 images | 352 x 704, 448 x 576 | ResNet-50, ResNet-101, ResNeXt-101 |
| STTR | KITTI, NYU Depth v2 | Monocular | N/A | 160 x 512, 228 x 314 | ResNet-101, DenseNet-161, SENet-154 |
| MonoIndoor | KITTI, Make3D | Monocular | 39,810 sequences, 400 images | N/A | Monodepth2 (resnet-101) [14] |
| Point-Fusion | KITTI | Monocular + Stereo | 200 stereo image pairs, 35,454 images | 256 x 512 | N/A |

Table 4: Summary of primary studies based on training data for depth estimation.

| Methods | Datasets | Training Supervision | Training Data Quantity | Image Size | Base Model |
|---|---|---|---|---|---|
| MDE-DM | NYU Depth v2 | N/A | N/A | 320 x 256 | ResNet-50, DenseNet-161, SENet-154, MobileNetV2 |
| IEN | KITTI, NYU Depth v2 | Monocular | 39810 images, 249 scenes | 192 x 640, 228 x 304 | ResNet-50, SENet-154 |
| ACAN | KITTI, NYU Depth v2 | Monocular | 22600 images, 284 scenes, 50,688 RGB and depth pairs | 512 x 256, 320 x 240 | ResNet, SENet, ChebNet |
| AG-GAN | KITTI, Make3D | N/A | 39,810 images, 400 images | 640 x 192, 1704 x 852 | ResNet-18, DepthNet |
| DBES-MDE | NYU Depth v2, KITTI, SUN RGB-D | Monocular | 50,000 images, 26,000 images, 5,050 images | 320 x 240, 704 x 352 | EfficientNet-B5 |
| DS-MDE | NYU Depth v2, KITTI | Monocular | 24,231 image pairs, 22600 images, | N/A | ResNet-101, Densenet-161 |
| StereoNet /SingleNet | NYU Depth v2, KITTI | Monocular | N/A, 22600 images | N/A | Vision Transformer (ViT) |
| Swin-Depth | NYU Depth v2, KITTI | Monocular | 24,231 image pairs, 23,488 images | 426 x 560, 352 x 704 | EfficienNet-B7, HRNet-64, Swin Transformer |
| FC-CRFs | KITTI, ETH3D, Middlebury dataset V3 | Stereo | N/A | N/A | U-Net |
| H-Net | KITTI, Flying things 3D, Middlebury dataset V3 | Stereo | N/A | 1242 x 375, 1024 x 436, N/A | Transformer |
| AMWR | EURoCMAV, NYU Depth V2, RGBD | Monocular | 05 sequences, 20,000 images, N/A | 512 × 256, 320 x 265, N/A | Monodepth2 |
| PCStereo | NYU Depth V2, KITTI | Multi-view | 60,000 images, 80,000 images | N/A | N/A |
| CORNet | NYU Depth V2 | Monocular | N/A | N/A | N/A |
| THE | KITTI, | Monocular | 39,810 | 128 x 416 | U-Net, ResNet-18 |
| LW-Net | NYU Depth v2, KITTI | Monocular | 249 sequences 22,600 images | N/A | ResNet-101 |
| CI-MDE | NYU Depth v2, KITTI | Monocular | 12,000 images, 22000 images | 256 x 352, 160 x 512 | Dilated-ResNet [47], ResNet-101, ResNet-50 |
| JAM-MDE | KITTI, HKUST-Drive | Stereo | 200 stereo image pairs | N/A | N/A |
| GMSN | KITTI, Cityscapes | Moncular | 22600 images, 22,973 images | 512 x 256 | CycleGAN, ResNet-50 |
| DDN-MDE | NYU Depth v2, KITTI | Moncular | 30,000 images, 39,810 images | N/A | EfficientNet-B3 |
| Multiscale-GCN | KITTTI, ETH3D | Stereo | N/A | 320 x 1000, 384 x 1000 | RAFT[35] |
| DFFN | Oxford RobotCar | N/A | 2 video sequences | 256 x 512 | CycleGAN |
| SABV-Depth | KITTI | Stereo | 22,600 images | N/A | U-Net |
| AMDE. | KITTI, Make3D | Monocular | N/A | 640 x 192 | ResNet-18 |
| MiniNet | KITTI, NYU Depth v2 | Monocular | 22,600 images 24,231 images | 224 x 896, 448 x 448 | Swin Transformer [8] |
| PVStereo | KITTI, NYU Depth v2, MatterPort3D | Monocular | 23,488 images, 249 scenes, 7,829 images | N/A, N/A, 1024 x 512 | Swin Transformer [21] |

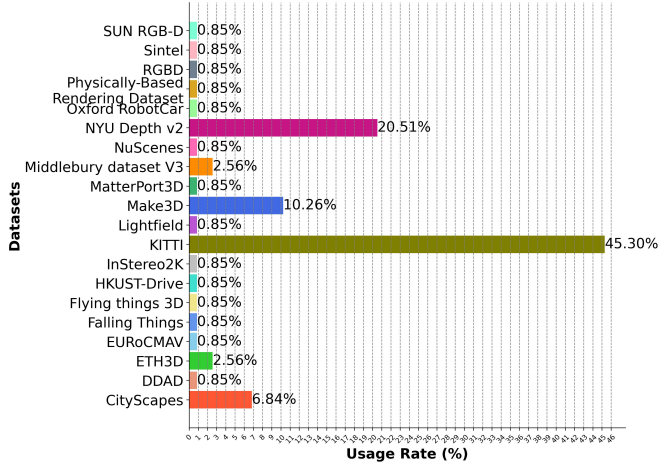

Figure 6: Usage rate of public datasets reported in the primary studies.

The primary studies have utilized different training supervision methods, as revealed by the usage rate of training supervision methods presented in Figure 7. In the context of DE, three types of training supervision methods have been reported, namely monocular-based, stereo-based, and monocular + stereo-based training supervision. Monocular-based training supervision involves using annotated depth maps or other depth-related cues to obtain ground truth for depth from monocular images, whereas stereo-based training supervision involves deriving ground truth depth from stereo image pairs. A hybrid approach, known as monocular + stereo-based training supervision, involves training the model using a combination of both monocular images and stereo-image pairs, which aims to leverage the advantages of both monocular and stereo depth cues for improved performance. Some studies have also reported the use of multi-view data for training supervision.

The most frequently used training supervision method was found to be monocular, accounting for 57.1% of the primary studies. Stereo-based training supervision was the second most popular, accounting for 34.7%, followed by monocular + stereo and multi-view, each accounting for 4.08%.

Regarding the amount of training data, most of the primary studies used pre-defined train/test splits as the training data, with no specific rule about how much data should be used to achieve effective and accurate results. The most used training data split was the KITTI eigen split [117], which comprised 22,600 training images. Similarly, training image size varied among the primary studies, with no defined criteria for selecting image size.

Additionally, 35 base models used to create the DE models were reported in the primary studies.

Table 5: Summary of primary studies based on training data for depth estimation.

| Methods | Datasets | Training Supervision | Training Data Quantity | Image Size | Base Model |
|---|---|---|---|---|---|
| ADAADepth | KITTI, Cityscapes | Stereo | 22,600 images | N/A | Monodepth2, U-Net, Resnet-18 |
| PA-Net | KITTI, Cityscapes, Make3D | Stereo | 22,600 images, 22,973 images, 400 images | N/A | ResNet-50 |
| SVDistNet | Middlebury dataset V3, ETH3D, KITTI, Sintel, Falling Things, InStereo2K | Stereo | 23 pairs of images, N/A 22,600 images, N/A, N/A, N/A | N/A | N/A |
| SI-MDE | KITTI | Stereo | 35,454 images | N/A | ResNet-50 |
| VADepth | KITTI | Monocular | 22,600 images | 640 x 192, or 416 x 128 | PackNet, DDVO, PoseNet |
| DE-MCS | KITTI | Monocular | 39,810 images | 640 x 192 | U-Net, ResNet-18 |
| SIPSNet | KITTI | Stereo | 35,454 images | N/A | N/A |
| Monodepth2 | KITTI, NYU Depth v2 | N/A | 23,488 images, 249 scenes | 416 x 128, 256 x 352 | ResNet-101 |
| VT | KITTI | Monocular | 22,600 images | N/A | N/A |
| HITNet | KITTI | Monocular | 39,810 images | N/A | DepthNet, PoseNet, DIFFNet [35], HRNet-18 |
| RAFT-Stereo | KITTI | Monocular | 39810 images | 192 x 640 | ResNet-18 |
| CREStereo | KITTI, Cityscapes | Monocular | 39,810 images, 69,731 images | 192 x 640, 128 x 416 | ResNet-18, ResNet-50, Transformer |
| SG-MDE | KITTI, NYU Depth v2 | Monocular | 22,600 images, 249 scenes | N/A | ShuffleNet |
| ACB | KITTTI, Make3D | Monocular | 39,810 images, 400 images | 192 x 640, 240 x 319 | ResNet-18, ResNet-50, U-Net |

The top five most-used base models were ResNet-50 (19.79%), ResNet-18 (10.42%), ResNet-101 (8.33%), U-Net (8.33%), and VGG-16 (5.21%), as shown by the usage rate of the base models in Figure 8.

3.4. RQ.4: Evaluation methods and metrics

The primary studies used 29 evaluation metrics to assess the performance of the DE. Figure 9 shows the usage rate of the evaluation metrics reported in the studies. The most consistent and prominent metrics were Root Mean Squared Error (RMSE) (18.18%), Accuracy (17.13%), Absolute Relative Difference (Abs Rel) (16.78%), Square Relative Difference (Sq Rel) (12.24%), RMSE log (9.09%), and log 10 (7.34%) among others.

RMSE is a metric defined as the square root of the average of squared differences between predicted $(x_i)$ and ground truth $(x_i^*)$ depth values for all pixels. Accuracy was reported in three different thresholds $\delta < 1.25$, $\delta < 1.25^2$, and $\delta < 1.25^3$, indicating the percentage of pixels for which the absolute relative error falls within a specified threshold. Abs Rel computes the average absolute relative difference between predicted and ground truth depth values for all pixels. Sq Rel measures the average squared relative difference between predicted and ground truth depth values for all pixels. These evaluation metrics are given below:

- RMSE $= \sqrt{\frac{1}{n}\sum_{i=1}^{n}(x_i - x_i^*)^2}$

- Accuracy with threshold ($\delta$) = % of $x_i$ s.t.

$$\max\left(\frac{x_i}{x_i^*}, \frac{x_i^*}{x_i}\right) \quad \delta < 1.25, \quad \delta < 1.25^2, \quad \delta < 1.25^3$$

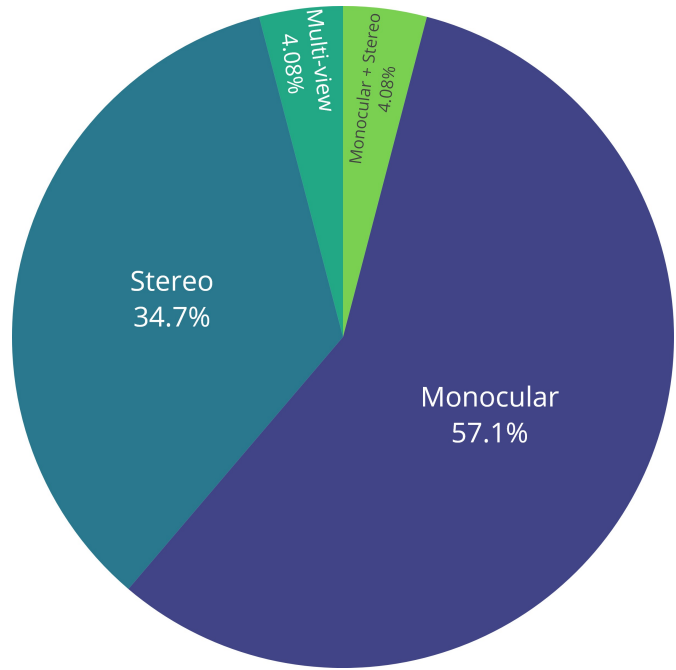

Figure 7: Usage rate of training supervision methods reported in the primary studies.

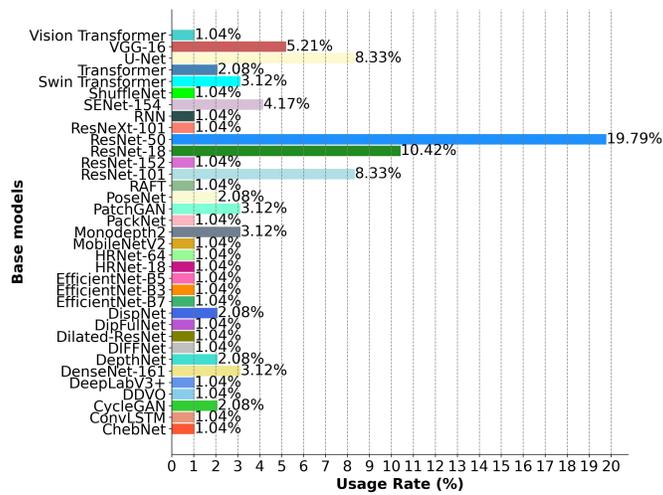

Figure 8: Usage rate of the base models reported in the primary studies.

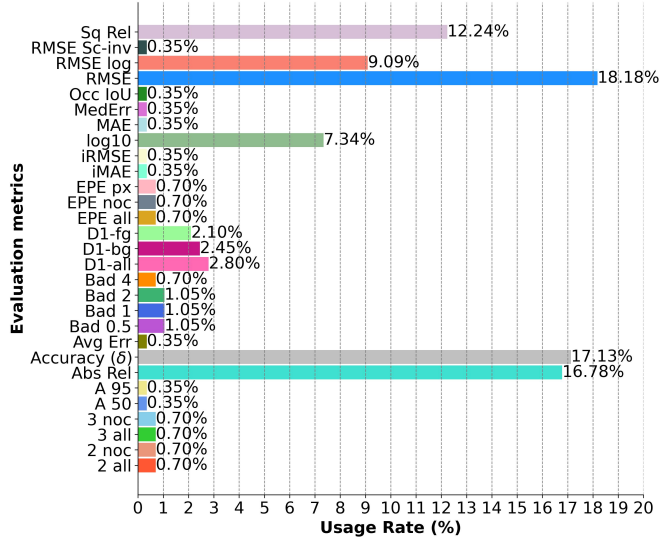

Figure 9: Usage rate of the evaluation metrics reported in the primary studies.

- Abs Rel $= 1/n \sum_i \left|\frac{x_i - x_i^*}{x_i}\right|$

- Sq Rel $= 1/n \sum_i \frac{||x_i - x_i^*||^2}{x_i}$

- RMSE (log) $= \sqrt{\frac{1}{n}\sum_i \left(\log(x_i) - \log(x_i^*)\right)^2}$

To provide a comprehensive analysis, the primary studies showcased their findings across various datasets including KITTI, NYU Depth V2, and Make3D. As such, this SLR study offers a detailed comparison of diverse DE methods across these datasets, as outlined in Tables 6, 7, and 8, utilising the performance metrics mentioned earlier. It is worth noting that the performance metrics that bear the symbol ↓ indicate that lower values are optimal, while those with the symbol ↑ suggest that higher values are preferable.

Table 6: Comparison of results of primary studies for the KITTI dataset.

| Reference | Methods | RMSE ↓ | RMSE log ↓ | RMSE Sc-inv ↓ | Abs Rel ↓ | Sq Rel ↓ | log10 ↓ | $\delta < 1.25$ ↑ | $\delta < 1.25^2$ ↑ | $\delta < 1.25^3$ ↑ |
|---|---|---|---|---|---|---|---|---|---|---|
| [31] | PHN | 4.082 | 0.164 | - | 0.136 | - | - | 0.864 | 0.966 | 0.989 |
| [32] | Deep CNN | 3.624 | 0.157 | 0.167 | 0.101 | 0.586 | 0.044 | 0.879 | 0.969 | 0.991 |
| [33] | T2Net | 4.717 | 0.245 | - | 0.169 | 1.23 | - | 0.769 | 0.912 | 0.965 |
| [34] | AdaDepth | 6.251 | 0.284 | - | 0.203 | 1.734 | - | 0.687 | 0.899 | 0.958 |
| [35] | SVSM | 3.266 | 0.167 | - | 0.09 | 0.499 | - | 0.902 | 0.968 | 0.986 |
| [36] | ConvNetD | 4.204 | 0.216 | - | 0.128 | 0.815 | - | 0.835 | 0.941 | 0.975 |
| [37] | PackNet | 3.485 | 0.121 | - | 0.078 | 0.42 | - | 0.931 | 0.986 | 0.996 |
| [38] | UnOS | 3.404 | 0.121 | - | 0.049 | 0.515 | - | 0.965 | 0.984 | 0.992 |
| [39] | Deep Siamese | 3.79 | 0.195 | - | 0.1069 | 0.6531 | - | 0.867 | 0.954 | 0.979 |
| [40] | DDC | 4.186 | 0.168 | - | 0.11 | 0.666 | - | 0.88 | 0.966 | 0.988 |
| [41] | monoResMatch | 4.714 | 0.199 | - | 0.096 | 0.673 | - | 0.864 | 0.954 | 0.979 |
| [42] | RNN for Depth | 1.698 | 0.11 | - | 0.077 | 0.205 | - | 0.941 | 0.99 | 0.998 |
| [73] | AMDE. | 4.405 | 0.181 | - | 0.107 | - | - | - | - | - |
| [82] | Monodepth2 | 4.63 | 0.193 | - | 0.106 | 0.806 | - | 0.876 | 0.958 | 0.98 |
| [65] | LW-Net | 4.483 | 0.192 | - | 0.115 | 0.853 | - | 0.874 | 0.959 | 0.981 |
| [66] | CI-MDE | 1.893 | 0.088 | - | 0.058 | 0.163 | - | 0.964 | 0.995 | 0.999 |
| [67] | JAM-MDE | 2.912 | 0.121 | - | 0.07 | 0.382 | - | 0.942 | 0.986 | 0.992 |
| [44] | SADDV-MDE | 4.699 | 0.185 | - | 0.106 | 0.861 | - | 0.889 | 0.962 | 0.982 |
| [69] | DDN-MDE | 4.327 | 0.171 | - | 0.096 | 0.655 | - | 0.893 | 0.963 | 0.983 |
| [70] | Multiscale-GCN | 4.256 | 0.177 | - | 0.097 | - | - | 0.918 | 0.97 | 0.985 |
| [74] | MiniNet | 4.067 | 0.205 | - | 0.135 | 0.839 | - | 0.838 | 0.947 | 0.978 |
| [45] | AdaBins | 2.36 | 0.088 | - | 0.058 | 0.19 | - | 0.964 | 0.995 | 0.999 |
| [46] | MAPUnet | 1.91 | 0.085 | - | 0.057 | 0.165 | - | 0.964 | 0.995 | 0.999 |
| [83] | VT | 2.573 | 0.092 | - | 0.062 | - | - | 0.959 | 0.995 | 0.999 |
| [47] | D-Net | 2.432 | 0.089 | - | 0.057 | 0.199 | - | 0.965 | 0.995 | 0.999 |
| [51] | MDE-DM | 4.726 | 0.167 | - | 0.109 | 0.842 | - | 0.897 | 0.972 | 0.989 |
| [52] | IEN | 3.4393 | 0.1718 | - | 0.1068 | - | - | 0.9004 | 0.976 | 0.9892 |
| [53] | ACAN | 3.509 | 0.118 | - | 0.075 | - | - | 0.93 | 0.985 | 0.996 |
| [54] | AG-GAN | 4.329 | 0.192 | - | 0.1196 | 0.889 | - | 0.865 | 0.943 | 0.989 |
| [55] | DBES-MDE | 3.325 | 0.116 | - | 0.074 | - | - | 0.933 | 0.989 | 0.997 |
| [57] | StereoNet /SingleNet | 4.392 | 0.185 | - | 0.094 | 0.681 | - | 0.892 | 0.962 | 0.981 |
| [76] | ADAADepth | 4.436 | 0.181 | - | 0.108 | 0.745 | - | 0.889 | 0.966 | 0.984 |
| [58] | Swin-Depth | 2.643 | 0.097 | - | 0.0654 | 0.232 | - | 0.957 | 0.994 | 0.999 |
| [59] | FC-CRFs | 2.129 | 0.079 | - | 0.052 | 0.155 | - | 0.974 | 0.997 | 0.999 |
| [60] | H-Net | 4.025 | 0.166 | - | 0.076 | 0.607 | - | 0.918 | 0.966 | 0.982 |
| [61] | AMWR | 3.931 | 0.201 | - | 0.115 | 0.712 | - | 0.857 | 0.951 | 0.978 |
| [87] | SG-MDE | 4.44 | 0.182 | - | 0.106 | 0.705 | - | 0.885 | 0.962 | 0.983 |
| [88] | ACB | 4.976 | 0.203 | - | 0.126 | 0.969 | - | 0.858 | 0.953 | 0.978 |
| [63] | CORNet | 2.603 | 0.1 | - | 0.06 | 0.265 | - | 0.959 | 0.994 | 0.998 |
| [78] | SVDistNet | 3.206 | 0.097 | - | 0.059 | 0.392 | - | 0.935 | 0.989 | 0.995 |
| [79] | SI-MDE | 4.087 | 0.167 | - | 0.09 | 0.597 | - | 0.912 | 0.97 | 0.985 |
| [80] | VADepth | 4.624 | 0.19 | - | 0.109 | 0.785 | - | 0.875 | 0.96 | 0.982 |
| [64] | THE | 4.317 | 0.174 | - | 0.095 | 0.696 | - | 0.902 | 0.965 | 0.983 |
| [72] | SABV-Depth | 3.458 | 0.158 | - | 0.107 | 0.817 | - | 0.892 | 0.959 | 0.991 |
| [3] | DE-MCS | 4.645 | 0.187 | - | 0.113 | 0.763 | - | 0.874 | 0.96 | 0.983 |

Based on the findings, the most successful methods for DE on KITTI dataset were FC-CRFs, MAPUnet, AdaBins, CI-MDE, and D-Net. Similarly, on the NYU Depth V2 dataset, the best methods were Point-Fusion, FC-CRFs, D-Net, Swin-Depth, and VT. Finally, on the Make3D dataset, the best approaches were PHN, AMDE, Deep CNN, ADAADepth, and DE-MCS.

### 3.5. RQ.5: Challenges and limitations in monocular and stereo DE

The challenges and limitations of monocular and stereo depth estimation (DE) techniques were thoroughly examined through the analysis of data from 59 primary studies. These challenges are crucial as they significantly affect the development and application of DE methods across various computer vision tasks. Below, we discuss the key challenges identified in these studies, their underlying causes, and the limitations they impose on DE systems.

Table 7: Comparison of results of primary studies for the NYU Depth V2 dataset.

| Reference | Methods | RMSE ↓ | RMSE log ↓ | RMSE Sc-inv ↓ | Abs Rel ↓ | Sq Rel ↓ | log10 ↓ | $\delta < 1.25$ ↑ | $\delta < 1.25^2$ ↑ | $\delta < 1.25^3$ ↑ |
|---|---|---|---|---|---|---|---|---|---|---|
| [31] | PHN | 0.501 | - | - | 0.144 | - | 0.184 | 0.835 | 0.962 | 0.992 |
| [32] | Deep CNN | 0.54 | 0.187 | 0.132 | 0.134 | 0.095 | 0.056 | 0.832 | 0.965 | 0.989 |
| [33] | T2Net | 0.915 | 0.305 | - | 0.257 | 0.281 | - | 0.54 | 0.832 | 0.948 |
| [34] | AdaDepth | 0.603 | - | - | 0.136 | - | 0.057 | 0.805 | 0.948 | 0.982 |
| [40] | DDC | 0.575 | 0.068 | - | 0.162 | - | - | 0.772 | 0.942 | 0.984 |
| [73] | AMDE. | 0.704 | - | - | 0.183 | - | 0.077 | 0.713 | 0.931 | 0.984 |
| [66] | CI-MDE | 0.388 | - | - | 0.111 | - | 0.047 | 0.878 | 0.981 | 0.995 |
| [67] | JAM-MDE | 0.523 | - | - | 0.113 | - | 0.049 | 0.872 | 0.975 | 0.993 |
| [68] | GMSN | 0.523 | - | - | 0.115 | - | 0.05 | 0.866 | 0.975 | 0.993 |
| [69] | DDN-MDE | 0.519 | - | - | 0.115 | - | 0.049 | 0.871 | 0.975 | 0.993 |
| [70] | Multiscale-GCN | 0.554 | - | - | 0.118 | - | 0.053 | 0.861 | 0.973 | 0.992 |
| [45] | AdaBins | 0.364 | - | - | 0.103 | - | 0.044 | 0.903 | 0.984 | 0.997 |
| [46] | MAPUnet | 0.393 | 0.04 | - | 0.109 | - | - | 0.888 | 0.979 | 0.997 |
| [83] | VT | 0.357 | - | - | 0.11 | - | 0.045 | 0.904 | 0.988 | 0.998 |
| [47] | D-Net | 0.354 | - | - | 0.095 | - | 0.041 | 0.919 | 0.988 | 0.997 |
| [49] | MonoIndoor | 0.526 | - | - | 0.134 | - | - | 0.823 | 0.958 | 0.989 |
| [50] | Point-Fusion | 0.126 | - | - | 0.022 | - | - | 0.994 | 0.999 | 1 |
| [71] | DFFN | 0.525 | - | - | 0.115 | - | 0.05 | 0.864 | 0.975 | 0.993 |
| [52] | IEN | 0.4983 | - | - | 0.1366 | - | 0.0583 | 0.8258 | 0.967 | 0.9949 |
| [53] | ACAN | 0.483 | 0.161 | - | 0.135 | - | - | 0.831 | 0.966 | 0.993 |
| [55] | DBES-MDE | 0.421 | - | - | 0.119 | - | 0.051 | 0.861 | 0.975 | 0.995 |
| [58] | Swin-Depth | 0.354 | - | - | 0.1 | - | 0.042 | 0.909 | 0.986 | 0.997 |
| [59] | FC-CRFs | 0.334 | 0.119 | - | 0.095 | - | 0.045 | 0.922 | 0.992 | 0.998 |
| [63] | CORNet | 0.47 | 0.05 | - | 0.109 | - | - | 0.859 | 0.973 | 0.995 |
| [72] | SABV-Depth | 0.421 | - | - | 0.101 | - | - | 0.894 | 0.977 | 0.989 |

One of the significant issues reported in DE is the lack of an effective mechanism to preserve cross-border details in depth maps. These details, especially at object edges, are essential for maintaining high-resolution and sharp depth predictions. Without accurately preserving them, depth maps often exhibit blurred or inaccurate boundaries, which degrades performance in applications such as robotics and autonomous navigation [31]. Ensuring the preservation of cross-border details is critical for improving the overall accuracy of DE models.

Monocular depth estimation (MDE) is frequently described as an inherently ill-posed problem, meaning multiple possible 3D configurations can exist for the same 2D image. Unlike stereo methods, which rely on disparity cues from multiple views, MDE must infer depth from a single image without explicit depth cues. This results in significant depth ambiguity, where the model infers relationships between objects based on limited information, leading to accumulated errors and a wide semantic gap [32]. This ill-posed nature of monocular DE remains a persistent challenge requiring innovative solutions.

In deep learning-based DE methods, a major challenge arises from data imbalance caused by the perspective effect. Objects closer to the camera occupy more space in an image, resulting in an overrepresentation of small depth values (nearby objects) and fewer samples of large depth values (distant objects). This imbalance skews the learning process, causing models to perform well on nearby objects but poorly on distant ones. Additionally, rapid depth changes at object boundaries complicate the prediction pro-cess, as models struggle to accurately capture these variations compared to other dense prediction tasks like segmentation [33].

Table 8: Comparison of results of primary studies for the Make3D dataset.

| Reference | Methods | RMSE ↓ | Abs Rel ↓ | Sq Rel ↓ | log10 ↓ |
|---|---|---|---|---|---|
| [31] | PHN | 4.32 | 0.179 | - | 0.066 |
| [32] | Deep CNN | 4.7 | 0.199 | - | 0.069 |
| [33] | T2Net | - | 0.508 | 6.589 | 0.574 |
| [34] | AdaDepth | 11.567 | 0.647 | 12.341 | - |
| [39] | Deep Siamese | 8.789 | 0.406 | 4.766 | 0.183 |
| [40] | DDC | 7.671 | 0.246 | - | 0.11 |
| [73] | AMDE. | 6.054 | 0.153 | - | 0.066 |
| [82] | Monodepth2 | 7.417 | 0.322 | 3.589 | 0.163 |
| [65] | LW-Net | 7.39 | 0.377 | 3.842 | 0.169 |
| [44] | SADDV-MDE | 7.013 | 0.297 | 2.902 | 0.158 |
| [74] | MiniNet | 8.534 | 0.398 | 5.167 | 0.192 |
| [76] | ADAADepth | 6.869 | 0.289 | 2.552 | 0.155 |
| [61] | AMWR | 7.745 | 0.352 | 4.115 | 0.176 |
| [3] | DE-MCS | 6.239 | 0.294 | 2.163 | - |

Acquiring ground truth data for training supervised DE models presents another significant challenge. Collecting paired RGB images and depth maps requires specialized depth-sensing equipment such as LiDAR or Kinect, which are both costly and complex to deploy. Even when obtained, ground truth datasets often contain noisy artifacts, particularly in reflective or dark environments, which can degrade DE model performance [34]. To address this data scarcity, synthetic datasets are frequently used, but they suffer from poor generalization. Synthetic images, while visually appealing, lack the textures, noise, and complexities of real-world environments, causing models trained on synthetic data to perform poorly when applied to real-world scenarios [33]. Additionally, Depth sensors like radar, LiDAR, sonar, structured light, and stereo cameras are widely used for data collection in industrial applications. However, these sensors have inherent limitations, including noise, sparse outputs, low resolution, and matching errors in stereo setups [47] [34]. These limitations introduce variability in depth data, reducing model accuracy and generalization in real-world settings.

Supervised DE models also rely heavily on high-level semantic information to link visual cues to depth. This often requires learning relationships between object recognition and DE, which is particularly difficult in unconstrained environments like outdoor scenes, where lighting, object sizes, and scene layouts vary widely. Learning such semantic information is computationally expensive and requires large, diverse datasets, making it challenging even with advanced loss functions designed to capture depth relationships [35].

Another critical challenge in DE is scale ambiguity, where objects at different distances may appear similarly sized in 2D images. To overcome this, DE models must capture long-range context information—the ability to infer relationships between distant parts of an image. Small, localized patches of an image often lack sufficient depth cues, so broader scene context must be incorporated for accurate depth predictions. Techniques like atrous spatial pyramid pooling and serialized layers extend the receptive field to capture multiscale objects. However, issues like discretized dilation, used to improve receptive fields while maintaining resolution, can lead to grid artifacts—undesired patterns or distortions in depth maps [53].

While stereo methods are often preferred for their ability to estimate depth through disparity between two images, monocular DE is appealing due to its simplicity and lack of hardware dependencies. Monocular DE leverages prior knowledge, such as monocular cues (texture gradients, occlusion, motion parallax), allowing for efficient 2D-to-3D problem-solving, and demonstrates potential in real-world scenarios. MDE eliminates the need for stereo rigs but still faces persistent challenges, including its ill-posed nature and difficulties in handling depth variations in complex environments [34]. Conversely, recent stereo-matching methods have improved disparity estimation in active stereo systems while maintaining accuracy. However, integrating such techniques into neural network-based stereo-matching systems for passive stereo-depth sensing remains an open question [84]. The choice between monocular and stereo methods often depends on the trade-offs between cost, accuracy, and deployment feasibility.

Monocular DE was found to be useful in medical imaging, particularly in procedures like endoscopy, where a single image is used to estimate depth. However, even in this domain, monocular DE encounters challenges like image artifacts during depth synthesis, reducing the accuracy and reliability of depth maps for medical diagnoses. These issues must be addressed to make DE systems viable in sensitive medical applications, such as anatomical reconstruction [3].

Although good results have been achieved with DE, challenges such as the inability to estimate the depth of certain objects, like windows, and the presence of artifacts persist [46]. Furthermore, the high sparsity of data highlights the need for further refinement and optimization in DE methods. Convolutional backbones are essential for extracting features in DE, but downsampling—the process of reducing image or feature map resolution in the deeper stages of the model—presents challenges for dense prediction tasks like depth estimation. Downsampling can result in a loss of fine details that are critical for accurate depth predictions. Therefore, careful consideration of architecture design is crucial to maintaining feature resolution throughout the network to improve depth prediction accuracy [83].

DE from videos presents unique challenges, as current models often fail to fully leverage temporal information—the movement of objects over time. This results in inaccuracies in motion estimation and camera pose changes [42]. Self-supervised video-based DE methods offer a promising alternative, but they require sophisticated network architectures and carry significant computational demands [44].

Furthermore, DE models were reported to often struggle to capture fine details and accurately estimate the depth of thin structures in images. This issue is particularly pronounced in higher-resolution images, where small details like wires or branches can be missed or misinterpreted. Real-world stereo image pairs further complicate this, as variations in camera characteristics and imperfect rectification introduce disparities and inaccuracies in depth predictions [86].

Generative Adversarial Networks (GANs) have been used to generate realistic depth maps from input images. In GANs, the discriminator refines predictions by distinguishing between real and generated depth maps, thereby improving accuracy. However, training GANs is challenging due to the complexity of balancing the generator and discriminator networks. Additionally, GANs often struggle to recover fine details and thin structures, limiting their application in high-accuracy DE tasks [43] [54].

Another challenge in both traditional and deep learning-based DE methods is non-Lambertian surfaces, such as shiny or metallic objects, which reflect light in unpredictable ways. These surfaces disrupt DE because standard reflection models fail. While some deep learning approaches have incorporated reflection-handling techniques, this remains an ongoing issue requiring further research [36]. Future research into deep learning solutions for non-Lambertian surfaces represents a promising direction.

Stereo depth estimation was also reported to face challenges in regions with non-textured or repetitive patterns, where it is difficult to find corresponding points between two images. This problem complicates DE in environments like walls or repetitive surfaces (e.g., roads), where stereo-matching algorithms typically fail. Addressing these disparities is essential for improving accuracy in real-world scenes [86].

Finally, the generalization of DE models remains a significant challenge. Existing benchmarks are often too small or limited in scope to adequately test how well models generalize to real-world environments. Larger, more varied datasets are needed to evaluate model performance across different environments and lighting conditions. Cross-domain generalization—the ability of a model trained on one type of data to perform well on another—remains particularly difficult and requires further research and experimentation [48].

While this review focused on answering the five selected research questions, there are several additional technical aspects of DE that warrant further consideration. One important development in self-supervised DE, as demonstrated by Monodepth2, is the integration of a pose estimation network between consecutive frames [82]. This allows the model to estimate relative camera motion, which enhances depth prediction by utilizing temporal information from video sequences. Such advances represent a significant step forward in self-supervised learning for DE.

Modern DE approaches often face challenges with discontinuous objects (e.g., object edges) and uncertainty in predictions. Techniques like edge-aware loss functions address these discontinuities, while methods like uncertainty estimation and confidence maps help quantify and manage uncertainty in depth predictions [89]. Additionally, implicit representations, such as Neural Radiance Fields (NeRFs), have gained attention for their ability to represent continuous volumetric scenes, offering new possibilities for DE. NeRFs are particularly exciting for their ability to capture fine-grained scene details, marking a significant advancement in the field.

The challenges and limitations of monocular and stereo DE methods require a holistic approach that considers aspects of data acquisition, training, model architecture, and application scenarios. Advancements in learning-based methods, the consideration of temporal information, efficient network architectures, and careful handling of real-world complexities are all necessary for overcoming these challenges. More research to fix these challenges is essential to unlocking the full potential of monocular and stereo DE in diverse applications.

## 4. Conclusion

The SLR study examines the application of DL techniques for DE. Accurate DE is of paramount importance for a wide range of applications, such as augmented reality and autonomous navigation, and this area has witnessed significant advancements with the aid of DL techniques.

The findings of this review underscore the immense potential of DL in DE tasks, surpassing traditional methods in both accuracy and efficiency. However, despite recent progress in learning-based methods, several challenges persist, such as data imbalance, rapid changes in depth values, and the need for long-range context. Obtaining paired training data remains a high-effort, high-cost endeavour, while ground truth acquisition is prone to noise and inaccuracies, limiting the effectiveness of supervised learning approaches.

Furthermore, the reliance on high-level semantic information poses challenges in establishing accurate depth relationships. Addressing these challenges is critical for practical navigation in real scenarios, where modelling scene dynamics and occlusion explicitly is essential. Although monocular DE offers several advantages over stereo setups, such as not requiring specialized equipment, the ill-posed nature of the problem persists, hindering accurate depth inference. Exploring alternate approaches, such as leveraging temporal information in monocular or stereo videos, may provide new avenues for improvement.

Finally, it should be noted that this review study only considers works published between 2018 and November 2023, selected from significant research databases using the criteria outlined in this study, while articles from other databases were not included in this review.

**CRediT authorship contribution statement**

Conceptualization: A.R.; Data curation: A.R. and M.J.H.; Formal analysis: A.R.; Investigation: A.R.; Methodology: A.R.; Software: A.R.; Supervision: A.R.; Validation: A.R.; Visualization: A.R. and M.J.H.; Writing – original draft: A.R.; Writing - review & editing: A.R., A.P., and M.J.H.

**Conflict of Interest**

The authors declare that there is no conflict of interest.

**Acknowledgement**

The authors would like to thank Prof. Jinchang Ren for the motivation behind the work.

**Declaration of generative AI and AI-assisted technologies in the writing process**

During the preparation of this work, the author(s) used ChatGPT to improve the English language and to correct possible grammatical mistakes. After using this tool/service, the author(s) reviewed and edited the content as needed and take(s) full responsibility for the content of the publication.

**References**

[1] T.-Y. Lin, J.-G. Juang, Application of 3d point cloud map and image identification to mobile robot navigation, Measurement and control (London) 56 (5-6) (2023) 911–927. doi:10.1177/00202940221136242.

[2] M. F. Afshar, Z. Shirmohammadi, S. A. A. G. Ghahramani, A. Noorparvar, A. M. A. Hemmatyar, An efficient approach to monocular depth estimation for autonomous vehicle perception systems, Sustainability (Basel, Switzerland) 15 (11) (2023) 8897. `doi:10.3390/su15118897`.

[3] G. Li, X. Chi, X. Qu, Depth estimation based on monocular camera sensors in autonomous vehicles: A self-supervised learning approach, Automotive Innovation 6 (2) (2023) 268–280. `doi:10.1007/s42154-023-00223-6`.

[4] T. T. Cocias, S. M. Grigorescu, F. Moldoveanu, Multiple-superquadrics based object surface estimation for grasping in service robotics, IEEE, May 2012, pp. 1471–1477. `doi:10.1109/OPTIM.2012.6231780`.

[5] A. K. Kushwaha, S. M. Khatavkar, D. M. Biradar, P. A. Chougule, Depth Estimation and Navigation Route Planning for Mobile Robots Based on Stereo Camera, Vol. 472 of Cognitive Computing and Cyber Physical Systems, Springer Nature Switzerland, Switzerland, 2023, pp. 180–191. `doi:10.1007/978-3-031-28975-0_15`.

[6] S. C. Diamantas, A. Oikonomidis, R. M. Crowder, Depth estimation for autonomous robot navigation: A comparative approach, IEEE, Jul 2010, pp. 426–430. `doi:10.1109/IST.2010.5548483`.

[7] Z. Zhang, M. Xiong, H. Xiong, Monocular depth estimation for uav obstacle avoidance, IEEE, Dec 2019, pp. 43–47. `doi:10.1109/CCIOT48581.2019.8980350`.

[8] B. Cetinkaya, S. Kalkan, E. Akbas, Does depth estimation help object detection?, Image and vision computing 122 (2022) 104427. `doi:10.1016/j.imavis.2022.104427`.

[9] M. Kalia, N. Navab, T. Salcudean, A real-time interactive augmented reality depth estimation technique for surgical robotics, IEEE, May 2019, pp. 8291–8297. `doi:10.1109/ICRA.2019.8793610`.

[10] Y.-M. Tsai, Y.-L. Chang, L.-G. Chen, Block-based vanishing line and vanishing point detection for 3d scene reconstruction, IEEE, Dec 2006, pp. 586–589. `doi:10.1109/ISPACS.2006.364726`.

[11] C. Tang, C. Hou, Z. Song, Depth recovery and refinement from a single image using defocus cues, Journal of modern optics 62 (6) (2015) 441–448. `doi:10.1080/09500340.2014.967321`.

[12] D. Lowe, Distinctive image features from scale-invariant keypoints, International journal of computer vision 60 (2) (2004) 91–110.

[13] N. Friedman, D. Koller, Probabilistic graphical models: principles and techniques, The MIT Press, 2009.

[14] S. Z. Li, Markov random field modeling in image analysis, 3rd Edition, Springer, London, 2009.

[15] H. Bay, A. Ess, T. Tuytelaars, L. V. Gool, Speeded-up robust features (surf), Computer vision and image understanding 110 (3) (2008) 346–359. `doi:10.1016/j.cviu.2007.09.014`.

[16] T.-Y. Kuo, P.-C. Su, Y.-P. Kuan, Sift-guided multi-resolution video inpainting with innovative scheduling mechanism and irregular patch matching, Information sciences 373 (2016) 95–109. `doi:10.1016/j.ins.2016.08.091`.

[17] E. Goldman, J. Goldberger, Crf with deep class embedding for large scale classification, Computer vision and image understanding 191 (2020) 102865. `doi:10.1016/j.cviu.2019.102865`.

[18] J. Y. Bouguet, Pyramidal implementation of the lucas kanade feature tracker. (1999).

[19] H. Hirschmuller, Accurate and efficient stereo processing by semi-global matching and mutual information, Vol. 2, IEEE, 2005, pp. 807–814 vol. 2. `doi:10.1109/CVPR.2005.56`.

[20] Z. Anton, K. Oleg, C. Sergei, N. Anatoliy, S. Sergei, Numerical methods for solving the problem of calibrating a projective stereo pair camera, optimized for implementation on fpga, Procedia Computer Science 167 (2020) 2229–2235. `doi:10.1016/j.procs.2020.03.275`.

[21] F. Liu, S. Zhou, Y. Wang, G. Hou, Z. Sun, T. Tan, Binocular light-field: Imaging theory and occlusion-robust depth perception application, IEEE transactions on image processing 29 (2020) 1628–1640. `doi:10.1109/TIP.2019.2943019`.

[22] X. Wang, J. Sun, H. Qin, Y. Yuan, J. Yu, Y. Su, Z. Sun, Accurate unsupervised monocular depth estimation for ill-posed region, Frontiers in physics 10 (Jan 12, 2023). `doi:10.3389/fphy.2022.1115764`.

[23] Y. Ming, X. Meng, C. Fan, H. Yu, Deep learning for monocular depth estimation: A review, Neurocomputing (Amsterdam) 438 (2021) 14–33. `doi:10.1016/j.neucom.2020.12.089`.

[24] A. Masoumian, H. A. Rashwan, J. Cristiano, M. S. Asif, D. Puig, Monocular depth estimation using deep learning: A review, Sensors (Basel, Switzerland) 22 (14) (2022) 5353. `doi:10.3390/s22145353`.

[25] P. Vyas, C. Saxena, A. Badapanda, A. Goswami, Outdoor monocular depth estimation: A research review, Tech. rep., Cornell University Library, arXiv.org (May 3, 2022). `doi:10.48550/arxiv.2205.01399`.

[26] C. Wang, X. Cui, S. Zhao, K. Guo, Y. Wang, Y. Song, The application of deep learning in stereo matching and disparity estimation: A bibliometric review, Expert systems with applications 238 (2024) 122006. `doi:10.1016/j.eswa.2023.122006`.

[27] H. Laga, L. V. Jospin, F. Boussaid, M. Bennamoun, A survey on deep learning techniques for stereo-based depth estimation, IEEE transactions on pattern analysis and machine intelligence 44 (4) (2022) 1738–1764. `doi:10.1109/TPAMI.2020.3032602`.

[28] B. Kitchenham, S. Charters, Guidelines for performing systematic literature reviews in software engineering, Technical Report EBSE 2007-001. Keele University and Durham University Joint Report. (2007).

[29] J. Cohen, Weighted kappa: Nominal scale agreement provision for scaled disagreement or partial credit, Psychological bulletin 70 (4) (1968) 213–220. `doi:10.1037/h0026256`.

[30] B. Kitchenham, R. Pretorius, D. Budgen, O. P. Brereton, M. Turner, M. Niazi, S. Linkman, Systematic literature reviews in software engineering – a tertiary study, Information and software technology 52 (8) (2010) 792–805. `doi:10.1016/j.infsof.2010.03.006`.

[31] Z. Zhang, C. Xu, J. Yang, J. Gao, Z. Cui, Progressive hard-mining network for monocular depth estimation, IEEE transactions on image processing 27 (8) (2018) 3691–3702. `doi:10.1109/TIP.2018.2821979`.

[32] B. Li, Y. Dai, M. He, Monocular depth estimation with hierarchical fusion of dilated cnns and soft-weighted-sum inference, Pattern recognition 83 (2018) 328–339. `doi:10.1016/j.patcog.2018.05.029`.

[33] C. Zheng, T.-J. Cham, J. Cai, T2net: Synthetic-to-realistic translation for solving single-image depth estimation tasks, Tech. rep., Cornell University Library, arXiv.org (Aug 4, 2018). `doi:10.48550/arxiv.1808.01454`.

[34] J. N. Kundu, P. K. Uppala, A. Pahuja, R. V. Babu, Adadepth: Unsupervised content congruent adaptation for depth estimation, IEEE, Jun 2018, pp. 2656–2665. `doi:10.1109/CVPR.2018.00281`.

[35] Y. Luo, J. Ren, M. Lin, J. Pang, W. Sun, H. Li, L. Lin, Single view stereo matching, IEEE, Jun 2018, pp. 155–163. `doi:10.1109/CVPR.2018.00024`.

[36] H. Zhan, R. Garg, C. S. Weerasekera, K. Li, H. Agarwal, I. M. Reid, Unsupervised learning of monocular depth estimation and visual odometry with deep feature reconstruction, IEEE, Jun 2018, pp. 340–349. `doi:10.1109/CVPR.2018.00043`.

[37] V. Guizilini, R. Ambrus, S. Pillai, A. Raventos, A. Gaidon, 3d packing for self-supervised monocular depth estimation, IEEE, Piscataway, Jun 2020, pp. 2482–2491. `doi:10.1109/CVPR42600.2020.00256`.

[38] Y. Wang, P. Wang, Z. Yang, C. Luo, Y. Yang, W. Xu, Unos: Unified unsupervised optical-flow and stereo-depth estimation by watching videos, IEEE, Jun 2019, pp. 8063–8073. `doi:10.1109/CVPR.2019.00826`.

[39] M. Goldman, T. Hassner, S. Avidan, Learn stereo, infer mono: Siamese networks for self-supervised, monocular, depth estimation, IEEE, Jun 2019, pp. 2886–2895. `doi:10.1109/CVPRW.2019.00348`.

[40] S. Gur, L. Wolf, Single image depth estimation trained via depth from defocus cues, IEEE, Piscataway, Jun 2019, pp. 7675–7684. `doi:10.1109/CVPR.2019.00787`.

[41] F. Tosi, F. Aleotti, M. Poggi, S. Mattoccia, Learning monocular depth estimation infusing traditional stereo knowledge, IEEE, Jun 2019, pp. 9791–9801. `doi:10.1109/CVPR.2019.01003`.

[42] R. Wang, S. M. Pizer, J.-M. Frahm, Recurrent neural network for (un-)supervised learning of monocular video visual odometry and depth, IEEE, Piscataway, Jun 2019, pp. 5550–5559. `doi:10.1109/CVPR.2019.00570`.

[43] Z. Wu, X. Wu, X. Zhang, S. Wang, L. Ju, Spatial correspondence with generative adversarial network: Learning depth from monocular videos, IEEE, Oct 2019, pp. 7493–7503. `doi:10.1109/ICCV.2019.00759`.

[44] A. Johnston, G. Carneiro, Self-supervised monocular trained depth estimation using self-attention and discrete disparity volume, IEEE, Jun 2020, pp. 4755–4764. `doi:10.1109/CVPR42600.2020.00481`.

[45] S. F. Bhat, I. Alhashim, P. Wonka, Adabins: Depth estimation using adaptive bins, IEEE, Piscataway, Jun 2021, pp. 4008–4017. `doi:10.1109/CVPR46437.2021.00400`.

[46] Y. Yang, Y. Wang, C. Zhu, M. Zhu, H. Sun, T. Yan, Mixed-scale unet based on dense atrous pyramid for monocular depth estimation, IEEE access 9 (2021) 114070–114084. `doi:10.1109/ACCESS.2021.3104605`.

[47] J. L. Thompson, S. L. Phung, A. Bouzerdoum, D-net: A generalised and optimised deep network for monocular depth estimation, IEEE access 9 (2021) 134543–134555. `doi:10.1109/ACCESS.2021.3116380`.

[48] Z. Li, X. Liu, N. Drenkow, A. Ding, F. X. Creighton, R. H. Taylor, M. Unberath, Revisiting stereo depth estimation from a sequence-to-sequence perspective with transformers, IEEE, Piscataway, Oct 2021, pp. 6177–6186. `doi:10.1109/ICCV48922.2021.00614`.

[49] P. Ji, R. Li, B. Bhanu, Y. Xu, Monoindoor: Towards good practice of self-supervised monocular depth estimation for indoor environments, IEEE, Piscataway, Oct 2021, pp. 12767–12776. `doi:10.1109/ICCV48922.2021.01255`.

[50] L. Huynh, P. Nguyen, J. Matas, E. Rahtu, J. Heikkila, Boosting monocular depth estimation with lightweight 3d point fusion, IEEE, Piscataway, Oct 2021, pp. 12747–12756. `doi:10.1109/ICCV48922.2021.01253`.

[51] H. Wang, Y. Sun, Q. M. J. Wu, X. Lu, X. Wang, Z. Zhang, Self-supervised monocular depth estimation with direct methods, Neurocomputing (Amsterdam) 421 (2021) 340–348. `doi:10.1016/j.neucom.2020.10.025`.

[52] W. Su, H. Zhang, Q. Zhou, W. Yang, Z. Wang, Monocular depth estimation using information exchange network, IEEE transactions on intelligent transportation systems 22 (6) (2021) 3491–3503. `doi:10.1109/TITS.2020.3008991`.

[53] Y. Chen, H. Zhao, Z. Hu, J. Peng, Attention-based context aggregation network for monocular depth estimation, International journal of machine learning and cybernetics 12 (6) (2021) 1583–1596. `doi:10.1007/s13042-020-01251-y`.

[54] S. Bhattacharyya, J. Shen, S. Welch, C. Chen, Efficient unsupervised monocular depth estimation using attention guided generative adversarial network, Journal of real-time image processing 18 (4) (2021) 1357–1368. `doi:10.1007/s11554-021-01092-0`.

[55] U. Ali, B. Bayramli, T. Alsarhan, H. Lu, A lightweight network for monocular depth estimation with decoupled body and edge supervision, Image and vision computing 113 (2021) 104261. `doi:10.1016/j.imavis.2021.104261`.

[56] L. Liu, X. Song, M. Wang, Y. Liu, L. Zhang, Self-supervised monocular depth estimation for all day images using domain separation, IEEE, Piscataway, Oct 2021, pp. 12717–12726. `doi:10.1109/ICCV48922.2021.01250`.

[57] Z. Chen, X. Ye, W. Yang, Z. Xu, X. Tan, Z. Zou, E. Ding, X. Zhang, L. Huang, Revealing the reciprocal relations between self-supervised stereo and monocular depth estimation (Oct 2021) 15509–15518`doi:10.1109/ICCV48922.2021.01524`.

[58] Z. Cheng, Y. Zhang, C. Tang, Swin-depth: Using transformers and multi-scale fusion for monocular-based depth estimation, IEEE sensors journal 21 (23) (2021) 26912–26920. `doi:10.1109/JSEN.2021.3120753`.

[59] W. Yuan, X. Gu, Z. Dai, S. Zhu, P. Tan, Neural window fully-connected crfs for monocular depth estimation, IEEE, Piscataway, Jun 2022, pp. 3906–3915. `doi:10.1109/CVPR52688.2022.00389`.

[60] B. Huang, J.-Q. Zheng, S. Giannarou, D. S. Elson, H-net: Unsupervised attention-based stereo depth estimation leveraging epipolar geometry, IEEE, Piscataway, Jun 2022, pp. 4459–4466. `doi:10.1109/CVPRW56347.2022.00492`.

[61] C. Ling, X. Zhang, H. Chen, Unsupervised monocular depth estimation using attention and multi-warp reconstruction, IEEE transactions on multimedia 24 (2022) 2938–2949. `doi:10.1109/TMM.2021.3091308`.

[62] W. Chuah, R. Tennakoon, R. Hoseinnezhad, A. Bab-Hadiashar, Deep learning-based incorporation of planar constraints for robust stereo depth estimation in autonomous vehicle applications, IEEE transactions on intelligent transportation systems 23 (7) (2022) 6654–6665. `doi:10.1109/TITS.2021.3060001`.

[63] X. Meng, C. Fan, Y. Ming, H. Yu, Cornet: Context-based ordinal regression network for monocular depth estimation, IEEE transactions on circuits and systems for video technology 32 (7) (2022) 4841–4853. `doi:10.1109/TCSVT.2021.3128505`.

[64] S.-J. Hwang, S.-J. Park, J.-H. Baek, B. Kim, Self-supervised monocular depth estimation using hybrid transformer encoder, IEEE sensors journal 22 (19) (2022) 18762–18770. `doi:10.1109/JSEN.2022.3199265`.

[65] C. Feng, C. Zhang, Z. Chen, M. Li, H. Chen, B. Fan, Lw-net: A lightweight network for monocular depth estimation, IEEE access 8 (2020) 196287–196298. `doi:10.1109/ACCESS.2020.3034751`.

[66] D. Kim, S. Lee, J. Lee, J. Kim, Leveraging contextual information for monocular depth estimation, IEEE access 8 (2020) 147808–147817. `doi:10.1109/ACCESS.2020.3016008`.

[67] P. Liu, Z. Zhang, Z. Meng, N. Gao, Joint attention mechanisms for monocular depth estimation with multi-scale convolutions and adaptive weight adjustment, IEEE access 8 (2020) 184437–184450. `doi:10.1109/ACCESS.2020.3030097`.

[68] L. Lin, G. Huang, Y. Chen, L. Zhang, B. He, Efficient and high-quality monocular depth estimation via gated multi-scale network, IEEE access 8 (2020) 7709–7718. `doi:10.1109/ACCESS.2020.2964733`.

[69] J. Wang, G. Zhang, M. Yu, T. Xu, T. Luo, Attention-based dense decoding network for monocular depth estimation, IEEE access 8 (2020) 85802–85812. `doi:10.1109/ACCESS.2020.2990643`.

[70] J. Fu, J. Liang, Z. Wang, Monocular depth estimation based on multi-scale graph convolution networks, IEEE access 8 (2020) 997–1009. `doi:10.1109/ACCESS.2019.2961606`.

[71] X. Yang, Q. Chang, X. Liu, S. He, Y. Cui, Monocular depth estimation based on multi-scale depth map fusion, IEEE access 9 (2021) 67696–67705. `doi:10.1109/ACCESS.2021.3076346`.

[72] J. Wang, Y. Chen, Z. Dong, M. Gao, H. Lin, Q. Miao, Sabv-depth: A biologically inspired deep learning network for monocular depth estimation, Knowledge-based systems 263 (2023) 110301. `doi:10.1016/j.knosys.2023.110301`.

[73] R. Ji, K. Li, Y. Wang, X. Sun, F. Guo, X. Guo, Y. Wu, F. Huang, J. Luo, Semi-supervised adversarial monocular depth estimation, IEEE transactions on pattern analysis and machine intelligence 42 (10) (2020) 2410–2422. `doi:10.1109/TPAMI.2019.2936024`.

[74] J. Liu, Q. Li, R. Cao, W. Tang, G. Qiu, Mininet: An extremely lightweight convolutional neural network for real-time unsupervised monocular depth estimation, ISPRS journal of photogrammetry and remote sensing 166 (2020) 255–267. `doi:10.1016/j.isprsjprs.2020.06.004`.

[75] H. Wang, R. Fan, P. Cai, M. Liu, Pvstereo: Pyramid voting module for end-to-end self-supervised stereo matching, IEEE robotics and automation letters 6 (3) (2021) 4353–4360. `doi:10.1109/LRA.2021.3068108`.

[76] V. Kaushik, K. Jindgar, B. Lall, Adaadepth: Adapting data augmentation and attention for self-supervised monocular depth estimation, IEEE robotics and automation letters 6 (4) (2021) 7791–7798. `doi:10.1109/LRA.2021.3101049`.

[77] X. Yu, J. Gu, Z. Huang, Z. Zhang, Parallax attention stereo matching network based on the improved group-wise correlation stereo network, PloS one 17 (2) (2022) e0263735. `doi:10.1371/journal.pone.0263735`.

[78] V. R. Kumar, M. Klingner, S. Yogamani, M. Bach, S. Milz, T. Fingscheidt, P. Mader, Svdistnet: Self-supervised near-field distance estimation on surround view fisheye cameras, IEEE transactions on intelligent transportation systems 23 (8) (2022) 10252–10261. `doi:10.1109/TITS.2021.3088950`.

[79] P. Jiang, W. Yang, X. Ye, X. Tan, M. Wu, Detaching and boosting: Dual engine for scale-invariant self-supervised monocular depth estimation, IEEE robotics and automation letters 7 (4) (2022) 1–8. `doi:10.1109/LRA.2022.3210877`.

[80] J. Xiang, Y. Wang, L. An, H. Liu, Z. Wang, J. Liu, Visual attention-based self-supervised absolute depth estimation using geometric priors in autonomous driving, IEEE robotics and automation letters 7 (4) (2022) 1–8. `doi:10.1109/LRA.2022.3210298`.

[81] Y. Tian, Y. Du, Q. Zhang, J. Cheng, Z. Yang, Depth estimation for advancing intelligent transport systems based on self-improving pyramid stereo network, IET intelligent transport systems 14 (5) (2020) 338–345. `doi:10.1049/iet-its.2019.0462`.

[82] C. Godard, O. M. Aodha, M. Firman, G. Brostow, Digging into self-supervised monocular depth estimation, IEEE, Piscataway, Oct 2019, pp. 3827–3837. `doi:10.1109/ICCV.2019.00393`.

[83] R. Ranftl, A. Bochkovskiy, V. Koltun, Vision transformers for dense prediction, IEEE, Piscataway, Oct 2021, pp. 12159–12168. `doi:10.1109/ICCV48922.2021.01196`.

[84] V. Tankovich, C. Hane, Y. Zhang, A. Kowdle, S. Fanello, S. Bouaziz, Hitnet: Hierarchical iterative

tile refinement network for real-time stereo matching, IEEE, Piscataway, Jun 2021, pp. 14357–14367. `doi:10.1109/CVPR46437.2021.01413`.

[85] L. Lipson, Z. Teed, J. Deng, Raft-stereo: Multilevel recurrent field transforms for stereo matching, IEEE, Piscataway, Dec 2021, pp. 218–227. `doi:10.1109/3DV53792.2021.00032`.

[86] J. Li, P. Wang, P. Xiong, T. Cai, Z. Yan, L. Yang, J. Liu, H. Fan, S. Liu, Practical stereo matching via cascaded recurrent network with adaptive correlation, IEEE, Piscataway, Jun 2022, pp. 16242–16251. `doi:10.1109/CVPR52688.2022.01578`.

[87] X. Lu, H. Sun, X. Wang, Z. Zhang, H. Wang, Semantically guided self-supervised monocular depth estimation, IET image processing 16 (5) (2022) 1293–1304. `doi:10.1049/ipr2.12409`.

[88] L. Hu, H. Zhang, Z. Wang, C. Huang, C. Zhang, Self-supervised monocular depth estimation via asymmetric convolution block, IET Cyber-Systems and Robotics 4 (2) (2022) 131–138. `doi:10.1049/csy2.12051`.

[89] Z. Yang, P. Wang, W. Xu, L. Zhao, R. Nevatia, Unsupervised learning of geometry from videos with edge-aware depth-normal consistency, Proceedings of the ... AAAI Conference on Artificial Intelligence 32 (1) (Apr 27, 2018). `doi:10.1609/aaai.v32i1.12257`.

[90] V. Garg, A. Kalai, Supervising unsupervised learning, Tech. rep., Cornell University Library, arXiv.org (Sep 14, 2017). `doi:10.48550/arxiv.1709.05262`.

[91] D. Xu, W. Wang, H. Tang, H. Liu, N. Sebe, E. Ricci, Structured attention guided convolutional neural fields for monocular depth estimation, IEEE, Jun 2018, pp. 3917–3925. `doi:10.1109/CVPR.2018.00412`.

[92] C. Yan, L. Li, C. Zhang, B. Liu, Y. Zhang, Q. Dai, Cross-modality bridging and knowledge transferring for image understanding, IEEE transactions on multimedia 21 (10) (2019) 2675–2685. `doi:10.1109/TMM.2019.2903448`.

[93] J. Lokoč, T. Skopal, K. Schoeffmann, V. Mezaris, X. Li, S. Vrochidis, I. Patras, MSCANet: Adaptive Multi-scale Context Aggregation Network for Congested Crowd Counting, Vol. 12573 of MMM (2), Springer International Publishing AG, Switzerland, 2021, pp. 1–12. `doi:10.1007/978-3-030-67835-7_1`.

[94] P. W. X. C. R. Yang, Depth estimation via affinity learned with convolutional spatial propagation network (2018). `doi:https://doi.org/10.48550/arXiv.1808.00150`.

[95] C. Godard, O. M. Aodha, G. J. Brostow, Unsupervised monocular depth estimation with left-right consistency, IEEE, Jul 2017, pp. 6602–6611. `doi:10.1109/CVPR.2017.699`.

[96] D. Lin, G. Chen, D. Cohen-Or, P.-A. Heng, H. Huang, Cascaded feature network for semantic segmentation of rgb-d images, IEEE, Oct 2017, pp. 1320–1328. `doi:10.1109/ICCV.2017.147`.

[97] S. Zhao, M. Gong, H. Fu, D. Tao, Adaptive context-aware multi-modal network for depth completion, IEEE transactions on image processing 30 (2021) 5264–5276. `doi:10.1109/TIP.2021.3079821`.

[98] W. Guo, Z. Li, Y. Yang, Z. Wang, R. Taylor, M. Unberath, A. Yuille, Y. Li, Context-enhanced stereo transformer, Tech. rep., Cornell University Library, arXiv.org (Oct 21, 2022). `doi:10.48550/arxiv.2210.11719`.

[99] M. Cordts, M. Omran, S. Ramos, T. Rehfeld, M. Enzweiler, R. Benenson, U. Franke, S. Roth, B. Schiele, The cityscapes dataset for semantic urban scene understanding, IEEE, Jun 2016, pp. 3213–3223. `doi:10.1109/CVPR.2016.350`.

[100] T. Schops, J. L. Schonberger, S. Galliani, T. Sattler, K. Schindler, M. Pollefeys, A. Geiger, A multi-view stereo benchmark with high-resolution images and multi-camera videos, IEEE, Jul 2017, pp. 2538–2547. `doi:10.1109/CVPR.2017.272`.

[101] M. Burri, J. Nikolic, P. Gohl, T. Schneider, J. Rehder, S. Omari, M. W. Achtelik, R. Siegwart, The euroc micro aerial vehicle datasets, The International journal of robotics research 35 (10) (2016) 1157–1163. `doi:10.1177/0278364915620033`.

[102] J. Tremblay, T. To, S. Birchfield, Falling things: A synthetic dataset for 3d object detection and pose estimation, IEEE, Jun 2018, pp. 2119–21193. `doi:10.1109/CVPRW.2018.00275`.

[103] N. Mayer, E. Ilg, P. Hausser, P. Fischer, D. Cremers, A. Dosovitskiy, T. Brox, A large dataset to train convolutional networks for disparity, optical flow, and scene flow estimation, IEEE, Jun 2016, pp. 4040–4048. `doi:10.1109/CVPR.2016.438`.

[104] W. Bao, W. Wang, Y. Xu, Y. Guo, S. Hong, X. Zhang, Instereo2k: a large real dataset for stereo matching in indoor scenes, Science China. Information sciences 63 (11) (2020) 212101. `doi:10.1007/s11432-019-2803-x`.

[105] A. Geiger, P. Lenz, R. Urtasun, Are we ready for autonomous driving? the kitti vision benchmark suite, IEEE, Jun 2012, pp. 3354–3361. `doi:10.1109/CVPR.2012.6248074`.

[106] S.-H. Lai, V. Lepetit, K. Nishino, Y. Sato, A Dataset and Evaluation Methodology for Depth Estimation on 4D Light Fields, Vol. 10113 of ACCV (3), Springer International Publishing AG, Switzerland, 2017, pp. 19–34. `doi:10.1007/978-3-319-54187-7_2`.

[107] F. Liu, C. Shen, G. Lin, I. Reid, Learning depth from single monocular images using deep convolutional neural fields, IEEE transactions on pattern analysis and machine intelligence 38 (10) (2016) 2024–2039. `doi:10.1109/TPAMI.2015.2505283`.

[108] A. Chang, A. Dai, T. Funkhouser, M. Halber, M. Niebner, M. Savva, S. Song, A. Zeng, Y. Zhang, Matterport3d: Learning from rgb-d data in indoor environments, IEEE, Oct 2017, pp. 667–676. `doi:10.1109/3DV.2017.00081`.

[109] D. Scharstein, H. Hirschmüller, Y. Kitajima, G. Krathwohl, N. Nešić, X. Wang, P. Westling, High-Resolution Stereo Datasets with Subpixel-Accurate Ground Truth, GCPR, Springer International Publishing, Cham, 2014, pp. 31–42. `doi:10.1007/978-3-319-11752-2_3`.

[110] H. Caesar, V. Bankiti, A. H. Lang, S. Vora, V. E. Liong, Q. Xu, A. Krishnan, Y. Pan, G. Baldan, O. Beijbom, nuscenes: A multimodal dataset for autonomous driving, IEEE, Piscataway, Jun 2020, pp. 11618–11628. `doi:10.1109/CVPR42600.2020.01164`.

[111] N. Silberman, D. Hoiem, P. Kohli, R. Fergus, Indoor Segmentation and Support Inference from RGBD Images, Computer Vision – ECCV 2012, Springer Berlin Heidelberg, Berlin, Heidelberg, 2012, pp. 746–760. `doi:10.1007/978-3-642-33715-4_54`.

[112] W. Maddern, G. Pascoe, C. Linegar, P. Newman, 1 year, 1000 km: The oxford robotcar dataset, The International journal of robotics research 36 (1) (2017) 3–15. `doi:10.1177/0278364916679498`.

[113] Y. Zhang, S. Song, E. Yumer, M. Savva, J.-Y. Lee, H. Jin, T. Funkhouser, Physically-based rendering for indoor scene understanding using convolutional neural networks, IEEE, Jul 2017, pp. 5057–5065. `doi:10.1109/CVPR.2017.537`.

[114] K. Lai, L. Bo, D. Fox, Unsupervised feature learning for 3d scene labeling, IEEE, May 2014, pp. 3050–3057. `doi:10.1109/ICRA.2014.6907298`.

[115] D. J. Butler, J. Wulff, G. B. Stanley, M. J. Black, A Naturalistic Open Source Movie for Optical Flow Evaluation, Computer Vision – ECCV 2012, Springer Berlin Heidelberg, Berlin, Heidelberg, 2012, pp. 611–625. `doi:10.1007/978-3-642-33783-3_44`.

[116] S. Song, S. P. Lichtenberg, J. Xiao, Sun rgb-d: A rgb-d scene understanding benchmark suite, IEEE, Jun 1, 2015, pp. 567–576. `doi:10.1109/CVPR.2015.7298655`.

[117] K. Lu, C. Zeng, Y. Zeng, Self-supervised learning of monocular depth using quantized networks, Neurocomputing (Amsterdam) 488 (2022) 634–646. `doi:10.1016/j.neucom.2021.11.071`.